\documentclass[11pt]{article}
\usepackage{color}
\usepackage[margin=1in]{geometry}
\usepackage{amssymb}
\usepackage{amsmath}
\usepackage{amsthm}
\usepackage{authblk}
\usepackage{booktabs}
\usepackage{longtable}
\usepackage{float,graphicx}
\usepackage{cases}
\usepackage{enumerate}
\usepackage[colorlinks=true]{hyperref}

\newtheorem{theorem}{Theorem}
\newtheorem{lemma}{Lemma}
\newtheorem{definition}{Definition}
\newtheorem{example}{Example}
\newtheorem{property}{Property}
\newtheorem{corollary}{Corollary}

\newcommand{\R}{\mathbb{R}}
\newcommand{\itp}{\hat{\otimes}_\varepsilon}
\newcommand{\norm}[1]{\left\lVert#1\right\rVert}

\title{MIONet: Learning multiple-input operators via tensor product}
\author[1]{Pengzhan Jin}
\author[2]{Shuai Meng}
\author[2,*]{Lu Lu}
\affil[1]{School of Mathematical Sciences, Peking University, Beijing 100871, China}
\affil[2]{Department of Chemical and Biomolecular Engineering, University of Pennsylvania, Philadelphia, PA 19104, USA}
\affil[*]{Corresponding author. Email: lulu1@seas.upenn.edu}
\date{}

\begin{document}

\maketitle

\begin{abstract}
As an emerging paradigm in scientific machine learning, neural operators aim to learn operators, via neural networks, that map between infinite-dimensional function spaces. Several neural operators have been recently developed. However, all the existing neural operators are only designed to learn operators defined on a single Banach space, i.e., the input of the operator is a single function. Here, for the first time, we study the operator regression via neural networks for multiple-input operators defined on the product of Banach spaces. We first prove a universal approximation theorem of continuous multiple-input operators. We also provide detailed theoretical analysis including the approximation error, which provides a guidance of the design of the network architecture. Based on our theory and a low-rank approximation, we propose a novel neural operator, MIONet, to learn multiple-input operators. MIONet consists of several branch nets for encoding the input functions and a trunk net for encoding the domain of the output function. We demonstrate that MIONet can learn solution operators involving systems governed by ordinary and partial differential equations. In our computational examples, we also show that we can endow MIONet with prior knowledge of the underlying system, such as linearity and periodicity, to further improve the accuracy.
\end{abstract}

\section{Introduction}

Scientific machine learning (SciML) field grows rapidly in recent years, where deep learning techniques are developed and applied to solve problems in computational science and engineering \cite{karniadakis2021physics}. As an active area of research in SciML, different methods have been developed to solve ordinary and partial differential equations (ODEs and PDEs) by parameterizing the solutions via neural networks (NNs), such as physics-informed NNs (PINNs) \cite{zhang2019quantifying,lu2021physics,pang2019fpinns,raissi2019physics,yu2021gradient},  deep Ritz method \cite{weinan2018deep}, and deep Galerkin method \cite{sirignano2018dgm}. These methods have shown promising results in diverse applications, such as fluid mechanics \cite{raissi2020hidden}, optics \cite{chen2020physics}, systems biology \cite{yazdani2020systems,daneker2022systems}, and biomedicine \cite{kissas2020machine}. However, these methods solve one specific instance of the PDE, and one needs to train a new neural network given a new initial condition, boundary condition, or forcing term, which is computationally costly and time consuming.

Another approach is applying neural networks (called neural operators) to learn solution operators of PDEs, mapping from an input function $v$ (e.g., initial condition, boundary condition, or forcing term) to the PDE solution $u$. This regression for the solution operator $\mathcal{G}$ is formulated as
\begin{equation*}
    \mathcal{G}: X \to Y,\quad v \mapsto u,
\end{equation*}
where $X$ and $Y$ are two infinite-dimensional Banach spaces of functions and $u=\mathcal{G}(v)$. We aim to learn $\mathcal{G}$ via NNs from a training dataset, i.e., some pairs $(v,\mathcal{G}(v))\in X\times Y$. Once a neural operator is trained, obtaining a solution $\mathcal{G}(v)$ for a new instance of $v$ requires only a forward pass of the network. 

Several approaches of neural operators have been recently proposed such as deep operator network (DeepONet) \cite{lu2019deeponet,lu2021learning,lu2021comprehensive} and Fourier neural operator (FNO) \cite{li2020fourier,lu2021comprehensive}, graph kernel network \cite{li2020neural,you2022nonlocal}, and others \cite{nelsen2021random,bhattacharya2020model,trask2019gmls,patel2021physics}. Among these approaches, DeepONet has been applied and demonstrated good performance in diverse applications, such as high-speed boundary layer problems \cite{di2021deeponet}, multiphysics and multiscale problems of hypersonics \cite{mao2021deepm} and electroconvection \cite{cai2021deepm}, multiscale bubble growth dynamics \cite{lin2021operator,lin2021seamless}, fractional derivative operators \cite{lu2021learning}, stochastic differential equations \cite{lu2021learning}, solar-thermal system \cite{osorio2022forecasting}, and aortic dissection \cite{yin2022simulating}. Several extensions of DeepONet have also been developed, such as Bayesian DeepONet \cite{lin2021accelerated}, DeepONet with proper orthogonal decomposition (POD-DeepONet) \cite{lu2021comprehensive}, multiscale DeepONet \cite{liu2021multiscale}, neural operator with coupled attention \cite{kissas2022learning}, and physics-informed DeepONet \cite{wang2021learning,goswami2022physics}.

Despite the progress of neural operators in computational tests, our theoretical understanding is lagging behind. The current theoretical analysis of neural operators focuses on the approximation capability such as the universal approximation theorems for DeepONet \cite{chen1995universal,lu2021learning} and FNO \cite{kovachki2021neural}. Recent theoretical works show that DeepONet \cite{deng2021convergence,lanthaler2021error} and FNO \cite{kovachki2021universal} may break the curse of dimensionality (CoD) in some problems, and DeepONet can approximate the solution operators of elliptic PDEs with exponential accuracy \cite{marcati2021exponential}. These results demonstrate the efficient approximation capability of DeepONet and FNO. One main difference between DeepONet and FNO is that the theory of FNO requires that the input function $v$ and the output function $u$ should be defined on the same domain \cite{kovachki2021universal,kovachki2021neural}, but the theory of DeepONet does not have this restriction \cite{chen1995universal,lu2021learning,deng2021convergence,lanthaler2021error,marcati2021exponential}.

However, all the existing neural operators are only designed to learn operators defined on a single Banach space $X$, i.e., the input of the operator is a single function. The theory of universal approximation for operators has only been proved for operators defined on a single Banach space. We note that some theoretical work \cite{kovachki2021neural} allows that the input function to be a vector-valued function, i.e., the input could be $\mathbf{v} = (v_1, v_2, \dots)$, but it still requires that all components of the input function $v_i$ must be defined on the same domain. This limitation of the input space prohibits us to learn a wide range of useful operators, e.g., the PDE solution operator mapping from both the initial condition and boundary condition to the PDE solution, as the initial condition and boundary condition are defined on two different domains (the initial domain and the boundary domain, respectively).

To overcome this limitation, in this work, we first study theoretically the approximation theory of the operator regression for a multiple-input operator $\mathcal{G}$ defined on the product of Banach spaces:
\begin{equation*}
    \mathcal{G}:X_1\times X_2\times\cdots\times X_n\to Y,
\end{equation*}
where $X_1,X_2,\cdots,X_n$ are $n$ different input Banach spaces, and $Y$ is the output Banach space. For example, $X_1$ can be the function space of all initial conditions, $X_2$ can be the function space of all boundary conditions, and $X_3$ can be the function space of all forcing terms, etc. Based on our theory, we then propose a novel neural operator, MIONet, to learn multiple-input operators. We verify in our computational results that MIONet can learn solution operators involving systems governed by ODEs and PDEs. We also discuss how we can endow MIONet with prior knowledge of the underlying system, such as linearity and periodicity, to further improve the accuracy.

This paper is organized as follows. In Section \ref{sec:theory}, we prove the approximation theory for multiple-input operator regression. Then we propose MIONet based on the theory in Section \ref{sec:method}. Subsequently, we test MIONet on several problems of ODEs and PDEs in Section \ref{sec:results}. Finally, Section \ref{sec:conclusions} summarizes this work.

\section{Approximation theory}
\label{sec:theory}

Our goal is to learn a (typically nonlinear) operator mapping from a product of $n$ Banach spaces (input spaces) to another Banach space (output space). These spaces are typically infinite dimensional. We first define the main notations used through out this paper. Denote the space comprised of all the continuous maps mapping from a metric space $X$ to a metric space $Y$ as $C(X,Y)$, and define $C(X):=C(X,\R)$. Let $X_1,X_2,\cdots,X_n$ and $Y$ be $n+1$ Banach spaces, and $K_i \subset X_i$ ($i=1, \cdots, n$) is a compact set. Then we aim to learn a continuous operator
\begin{equation*}
    \mathcal{G}: K_1\times \cdots\times K_n \to Y,\quad (v_1, \cdots, v_n) \mapsto u,
\end{equation*}
where $v_i\in K_i$ and $u=\mathcal{G}(v_1,\cdots,v_n)$. Such $\mathcal{G}$ form the space $C(K_1\times \cdots\times K_n,Y)$, which is studied in this paper.

In this section, we prove the approximation theory of continuous multiple-input operators by first illustrating our basic idea using the example of multilinearity on finite-dimensional spaces in Section \ref{sec:multilinear_operator}. We then introduce the techniques of Schauder basis and canonical projection for infinite-dimensional spaces in Section \ref{sec:finite_coordinates}, based on which we present the main theory of nonlinear operators in Section \ref{sec:nonlinear_operators} with more detailed analysis in Section \ref{sec:analysis}. We also provide a view of the theory through the tensor product of Banach spaces in Section \ref{sec:tensor}. All the proofs can be found in Appendix \ref{app:proofs}.

\subsection{Multilinear operators defined on finite-dimensional Banach spaces}
\label{sec:multilinear_operator}

We first use a simple case to illustrate the main idea of our theoretical approach: multilinear operators defined on finite-dimensional Banach spaces. Specifically, we consider a multilinear operator $$\mathcal{G}:X_1\times\cdots\times X_n\to Y,$$
where $X_1,\cdots,X_n$ are Banach spaces of finite dimensions $d_1,\cdots,d_n$.

Let $\{\phi_j^i\}_{j=1}^{d_i}\subset X_i$ be a basis of $X_i$, and thus for each $v_i \in X_i$, there exists a coordinate representation:
$$v_i = \sum_{j=1}^{d_i} \alpha_j^i\phi_j^i$$
for some vector $\alpha_i = (\alpha_1^i, \alpha_2^i, \cdots, \alpha_{d_i}^i) \in \R^{d_i}$. Because $\mathcal{G}$ is multilinear, for any input $(v_1, \cdots, v_n)$,
\begin{equation*}
\mathcal{G}(v_1,\cdots,v_n) = \mathcal{G} \left( \sum_{j_1=1}^{d_1}\alpha_{j_1}^1\phi_{j_1}^1, \cdots, \sum_{j_n=1}^{d_n}\alpha_{j_n}^n\phi_{j_n}^n\right)
= \sum_{j_1=1}^{d_1} \cdots \sum_{j_n=1}^{d_n} \mathcal{G}\left(\phi_{j_1}^1,\cdots,\phi_{j_n}^n\right) \alpha_{j_1}^1\cdots\alpha_{j_n}^n,
\end{equation*}
where $u_{j_1\cdots j_n} = \mathcal{G}\left(\phi_{j_1}^1,\cdots,\phi_{j_n}^n\right) \in Y$ is the output of $\mathcal{G}$ for the input $\left(\phi_{j_1}^1,\cdots,\phi_{j_n}^n\right)$. For convenience and clarity, for $\mathbf{u} = (u_{j_1\cdots j_n})_{d_1\times\cdots\times d_n}\in Y^{d_1\times\cdots\times d_n}$, we use the notation $\mathbf{u}\langle\cdots\rangle$ to represent the multilinear map
\begin{equation*}
    \mathbf{u}\langle \alpha_1,\cdots,\alpha_n\rangle := \sum_{j_1=1}^{d_1} \cdots \sum_{j_n=1}^{d_n} u_{j_1\cdots j_n}\alpha_{j_1}^1\cdots\alpha_{j_n}^n.
\end{equation*}
Hence, a multilinear operator defined on finite-dimensional Banach spaces can be represented as:
\begin{equation} \label{eq:multilinear_operator}
    \mathcal{G}(v_1,\cdots,v_n) = \left(\mathcal{G} \left(\phi_{j_1}^1,\cdots,\phi_{j_n}^n\right) \right)_{d_1\times\cdots\times d_n} \langle\alpha_1,\cdots,\alpha_n\rangle.
\end{equation}

Next we discuss the main idea of the approximation theory of $\mathcal{G}$, i.e., how to construct a surrogate model $\tilde{\mathcal{G}}_\theta$ (parameterized by the parameters $\theta$) to approximate $\mathcal{G}$. We note that $\alpha_i$ in Eq.~\eqref{eq:multilinear_operator} can be computed directly for $v_i$, and thus to approximate $\mathcal{G}$, it is sufficient to approximate $\left(\mathcal{G} \left(\phi_{j_1}^1,\cdots,\phi_{j_n}^n\right) \right)_{d_1\times\cdots\times d_n}$.
We consider $Y=C(K)$ for a compact set $K\subset \R^d$, and then we can construct $\tilde{\mathcal{G}}_\theta$ as
\begin{equation*}
\begin{split}
    \tilde{\mathcal{G}}_\theta:&\R^{d_1}\times\cdots\times \R^{d_n}\to C(K), \\
    &(\alpha_1,\cdots,\alpha_n)\mapsto \tilde{f}_{\theta}\langle\alpha_1,\cdots,\alpha_n\rangle,
\end{split}
\end{equation*}
where $\tilde{f}_{\theta}\in C(K,\R^{d_1\times\cdots\times d_n})$ is a function class parameterized by parameters $\theta$. It is easy to show that $\tilde{\mathcal{G}}_\theta$ is multilinear and can approximate $\mathcal{G}$ arbitrarily well as long as $\tilde{f}_{\theta}$ approximates $\left(\mathcal{G} \left(\phi_{j_1}^1,\cdots,\phi_{j_n}^n\right) \right)_{d_1\times\cdots\times d_n}$ well, which can be achieved by choosing $\tilde{f}_{\theta}$ as neural networks.

\subsection{Schauder basis and canonical projections for infinite-dimensional spaces}
\label{sec:finite_coordinates}

To deal with infinite-dimensional spaces, we introduce the Schauder basis and canonical projections. We refer the reader to \cite{fabian2011banach} for more details.
\begin{definition}[Schauder basis]
Let $X$ be an infinite-dimensional normed linear space. A sequence $\{e_i\}_{i=1}^{\infty}$ in $X$ is called a Schauder basis of $X$, if for every $x\in X$ there is a unique sequence of scalars $\{a_i\}_{i=1}^{\infty}$, called the coordinates of $x$, such that $$x=\sum_{i=1}^{\infty}a_i e_i.$$
\end{definition}
We show two useful examples of Schauder basis as follows.
\begin{example} \label{exa:faber}
\textbf{Faber-Schauder basis of $C[0,1]$.} Given distinct points $\{t_i\}_{i=1}^{\infty}$ which is a dense subset in $[0,1]$ with $t_1=0$, $t_2=1$. Let $e_1(t)=1$, $e_2(t)=t$, and $e_{k+1}$ is chosen as an element, such that ${e_1,\cdots,e_k,e_{k+1}}$ is a basis of the $(k+1)$-dimensional space which consists of all the piecewise linear functions with grid points $\{t_i\}_{i=1}^{k+1}$.
\end{example}
\begin{example}
\textbf{Fourier basis of $L^2[0,1]$.} Any orthogonal basis in a separable Hilbert space is a Schauder basis.
\end{example}

We denote the coordinate functional of $e_i$ by $e_i^*$, and thus
\begin{equation*}
    x=\sum_{i=1}^{\infty}e_i^*(x)e_i,\quad \forall x\in X.
\end{equation*}
Then for a constant $n$, the canonical projection $P_n$ is defined as
\begin{equation*}
    P_n(x)=P_n \left(\sum_{i=1}^{\infty}e_i^*(x)e_i\right) = \sum_{i=1}^{n}e_i^*(x)e_i.
\end{equation*}
We have the following property for $P_n$, according to which, we can represent points in an infinite-dimensional Banach space by finite coordinates within a sufficiently small projection error.

\begin{property}[Canonical projection] \label{pro:projection}
Assume that $K$ is a compact set in a Banach space $X$ equipped with a Schauder basis and corresponding canonical projections $P_n$, then we have
\begin{equation*}
    \lim_{n\to\infty}\sup_{x\in K}\norm{x-P_n(x)}=0.\\
\end{equation*}
\end{property}

For convenience, we decompose the $P_n$ as
\begin{equation*}
    P_n=\psi_n\circ\varphi_n,
\end{equation*}
where $\varphi_n:X\to\R^n$ and $\psi_n:\R^n\to X$ are defined as
\begin{equation*}
    \varphi_n(x) = \left(e_1^*(x),\cdots,e_n^*(x)\right)^T,\quad\psi_n(\alpha_1,\cdots,\alpha_n)=\sum_{i=1}^{n}\alpha_ie_i.
\end{equation*}
The $\varphi_n(x)$ are essentially the truncated coordinates for $x$. Moreover, sometimes we can further replace $\{e_1,\cdots,e_n\}$ with an equivalent basis for the decomposition of $P_n$, i.e.,
\begin{equation*}
    \hat{\varphi}_n(x)=Q(e_1^*(x),\cdots,e_n^*(x))^T,\quad\hat{\psi}_n(\alpha_1,\cdots,\alpha_n)=(e_1,\cdots,e_n)Q^{-1}(\alpha_1,\cdots,\alpha_n)^T,
\end{equation*}
with a nonsingular matrix $Q\in\R^{n\times n}$. For example, when applying the Faber-Schauder basis (Example \ref{exa:faber}), instead of using the coordinates based on the sequence $\{e_i\}_{i=1}^\infty$, we use the function values evaluated at certain grid points as the coordinates, which is the same as the linear element basis in the finite element method.

\subsection{Main theorems: Approximation theory for multiple-input operators}
\label{sec:nonlinear_operators}

Here, we present the main approximation theorems in Theorem \ref{thm:approximation} and Corollary \ref{cor:equiform}.

\begin{theorem} \label{thm:approximation}
Suppose that $X_1,\cdots,X_n,Y$ are Banach spaces, $K_i\subset X_i$ are compact sets, and $X_i$ have a Schauder basis with canonical projections $P_q^i=\psi_q^i\circ\varphi_q^i$. Assume that $\mathcal{G}:K_1\times\cdots\times K_n\to Y$ is a continuous operator, then for any $\epsilon>0$, there exist positive integers $p_i$, $q_i$, continuous vector functions $\mathbf{g}_i\in C(\R^{q_i},\R^{p_i})$, and $\mathbf{u}=(u_{j_1\cdots j_n})\in Y^{p_1\times\cdots\times p_n}$, such that
\begin{equation}\label{eq:approximation}
    \sup_{v_i\in K_i}\norm{\mathcal{G}(v_1,\cdots,v_n)-\mathbf{u} \left\langle\mathbf{g}_1(\varphi_{q_1}^1(v_1)),\cdots,\mathbf{g}_n(\varphi_{q_n}^n(v_n))\right\rangle}<\epsilon.
\end{equation}
\end{theorem}

\begin{corollary}\label{cor:equiform}
The conclusion in Theorem \ref{thm:approximation} can also be expressed in the following equivalent forms.
\begin{enumerate}[(i)]
    \item There exist positive integers $p_i,q_i,r$, continuous vector functions $\mathbf{g}_i\in C(\R^{q_i},\R^{p_i})$, $\mathbf{u}\in Y^r$, and $W\in\R^{p_1\times\cdots\times p_n\times r}$, such that
    \begin{equation}\label{eq:symm}
        \sup_{v_i\in K_i}\norm{\mathcal{G}(v_1,\cdots,v_n) - W \left\langle\mathbf{g}_1(\varphi_{q_1}^1(v_1)),\cdots,\mathbf{g}_n(\varphi_{q_n}^n(v_n)),\mathbf{u} \right\rangle}<\epsilon.
    \end{equation}
    If $\{e_i\}$ is a Schauder basis for $Y$, we can further have $\mathbf{u}=(e_1,e_2,\cdots,e_r)^T$.
    \item There exist positive integers $p,q_i$, continuous functions $g_{j}^i\in C(\R^{q_{i}})$, $u_j\in Y$, such that
\begin{equation}\label{eq:elementwise}
\sup_{v_i\in K_i}\norm{\mathcal{G}(v_1,\cdots,v_n)-\sum_{j=1}^{p}g_j^1(\varphi_{q_1}^1(v_1))\cdots g_j^n(\varphi_{q_n}^n(v_n))\cdot u_j}<\epsilon.
\end{equation}
\end{enumerate}
\end{corollary}

The relations between these three results are as follows. We first prove  \eqref{eq:approximation}; in (\ref{eq:symm}), we treat $\mathbf{g}_i$ and $\mathbf{u}$ in \eqref{eq:approximation} symmetrically and combine them via a tensor; and (\ref{eq:elementwise}) is simply a summation of products. In fact, when $Y$ is a space of continuous function approximated by fully-connected neural networks (FNNs), (\ref{eq:approximation}) and (\ref{eq:symm}) are technically equivalent, since $W$ can be regarded as the final linear output layer of the FNN for approximating $\mathbf{u}$. Therefore, we design two architectures in Section \ref{sec:method}, one based on (\ref{eq:approximation})/(\ref{eq:symm}) and the other based on (\ref{eq:elementwise}).

Next we show two special cases of $n=1$ based on theory above. In Example \ref{exa:UAT_faber}, we choose the Faber-Schauder basis as a Schauder basis. In Example \ref{exa:deeponet}, we have the universal approximation theorem for DeepONets.
\begin{example} \label{exa:UAT_faber}
Assume that $K$ is a compact set in $C[0,1]$, $\mathcal{G}:K\to C[0,1]$ is a continuous operator, then for any $\epsilon>0$, there exist positive integers $q,r$ and a continuous map $\mathbf{f}:\R^q\to\R^r$, such that
\begin{equation*}
    \norm{\mathcal{G}(v)-\sum_{i=1}^{r} f_i \left(v(\frac{0}{q-1}),v(\frac{1}{q-1}),\cdots,v(\frac{q-1}{q-1})\right) \cdot e_i}<\epsilon
\end{equation*}
holds for all $v\in K$, where $\mathbf{f}=(f_i)$. $\{e_i\}_{i=1}^r$ are chosen as the piecewise linear functions with grid points $\frac{j}{r-1}$, and $e_i(\frac{j-1}{r-1})=\delta_{ij}$. In fact, $f_i$ denotes the values of $\mathcal{G}(v)$ at $\frac{i-1}{r-1}$. This example is a direct conclusion of Example \ref{exa:faber} and Corollary \ref{cor:equiform}(i).
\end{example}

\begin{example}[DeepONet]
\label{exa:deeponet}

As a special case, for $n=1$ in Theorem \ref{thm:approximation}, we obtain the universal approximation theorem for DeepONet (Theorem 2 in \cite{lu2021learning}).
\end{example}

\subsection{Detailed analysis}
\label{sec:analysis}

We first introduce Lemma \ref{lem:functional} and Theorem \ref{thm:separate}, which are used to prove the main theorems in Section \ref{sec:nonlinear_operators}.

\begin{lemma} \label{lem:functional}
Suppose that $X_1,\cdots,X_n,Y$ are Banach spaces, $K_i\subset X_i$ are compact sets. Assume that $\mathcal{G}:K_1\times\cdots\times K_n\to Y$ is a continuous operator, then for any $\epsilon>0$, there exist positive integers $p_i$, continuous vector functionals $\hat{\mathbf{g}}_i\in C(X_i,\R^{p_i})$, and $\mathbf{u}\in Y^{p_1\times p_2\times\cdots\times p_n}$, such that
\begin{equation*}
    \sup_{v_i\in K_i}\norm{\mathcal{G}(v_1,\cdots,v_n)-\mathbf{u}\langle\hat{\mathbf{g}}_1(v_1),\cdots,\hat{\mathbf{g}}_n(v_n)\rangle}<\epsilon.
\end{equation*}
\end{lemma}

Lemma \ref{lem:functional} gives the approximation theory in the original infinite-dimensional Banach spaces. Next we extend to the following result.

\begin{theorem} \label{thm:separate}
Suppose that $X_1,\cdots,X_n,Y$ are Banach spaces, $K_i\subset X_i$ are compact sets, $X_i$ have a Schauder basis with canonical projections $P_q^i$. Assume that $\mathcal{G}:K_1\times K_2\times\cdots\times K_n\to Y$ is a continuous operator, then for any $\epsilon>0$, there exist positive integers $p_i$, continuous vector functionals $\hat{\mathbf{g}}_i\in C(X_i,\R^{p_i})$, and $\mathbf{u}\in Y^{p_1\times p_2\times\cdots\times p_n}$, such that
\begin{equation}\label{eq:separate}
    \sup_{v_i\in K_i}\norm{\mathcal{G}(v_1,\cdots,v_n)-\mathbf{u}\langle\hat{\mathbf{g}}_1(P_{q_1}^1(v_1)),\cdots,\hat{\mathbf{g}}_n(P_{q_n}^n(v_n))\rangle}<\epsilon+M\sum_{i=1}^n L_i^\epsilon(q_i)
\end{equation}
holds for arbitrary positive integers $q_i$, where
\begin{equation*}
L_i^\epsilon(q_i)=\sup_{v_i\in K_i}\norm{\hat{\mathbf{g}}_i\circ P_{q_i}^i(v_i)-\hat{\mathbf{g}}_i(v_i)}_1,\quad M=\max_{v_i\in K_i}\norm{\mathcal{G}(v_1,\cdots,v_n)}.
\end{equation*}
Note that $L_i^\epsilon(q_i)\to 0$ as $q_i\to\infty$.
\end{theorem}

Theorem \ref{thm:approximation} can be derived immediately from Theorem \ref{thm:separate} as long as we choose sufficiently large $q_i$ which makes $L_i^\epsilon$ small enough. In Eq.~\eqref{eq:separate}, the first part of error ``$\epsilon$'' is due to operator approximation, while the second part of error ``$M\sum_{i=1}^{n}L_i^\epsilon(q_i)$'' is due to the projection to finite-dimensional space. We note that $L_i^\epsilon$ depends on $\epsilon$, and thus when $\epsilon$ is small which makes $L_i^\epsilon$ converge slower, a large value of $q_i$ is needed.

Next we show further analysis of these results, which also provides a guidance of the design of the network architectures in Section \ref{sec:method}.

\begin{corollary}[Effect of a bias] \label{cor:bias}
In Theorem \ref{thm:separate}, if $Y=C(K_0)$ for compact $K_0$ in a Banach space $X_0$, we take an additional bias $b\in\R$, such that Eq.~(\ref{eq:separate}) becomes
\begin{equation*}
\sup_{v_i\in K_i,y\in K_0}|\mathcal{G}(v_1,\cdots,v_n)(y) - \mathbf{f}(y) \left\langle\hat{\mathbf{g}}_1(P_{q_1}^1(v_1)),\cdots,\hat{\mathbf{g}}_n(P_{q_n}^n(v_n))\right\rangle - b|<\epsilon+M\sum_{i=1}^n L_i^\epsilon(q_i),
\end{equation*}
where $\mathbf{f}\in C(K_0,\R^{p_1\times p_2\times\cdots\times p_n})$ and
\begin{equation*}
    M=\frac{1}{2}\left(\max_{v_i\in K_i,y\in K_0}\mathcal{G}(v_1,\cdots,v_n)(y)-\min_{v_i\in K_i,y\in K_0}\mathcal{G}(v_1,\cdots,v_n)(y)\right).
\end{equation*}
\end{corollary}

Corollary \ref{cor:bias} suggests us to add a bias, which makes the constant $M$ smaller and thus decreases the error. In addition, we explore more characteristics of Theorem \ref{thm:approximation} for learning multiple operators.

\begin{corollary}[Approximation theory for multiple operators] \label{cor:multiple_outputs}
\label{cor:multi_output}
Suppose that $X_1,\cdots,X_n,Y_1,\cdots,Y_m$ are Banach spaces, $K_i\subset X_i$ are compact sets, $X_i$ have a Schauder basis with canonical projections $P_q^i=\psi_q^i\circ\varphi_q^i$. Assume that $\mathcal{G}_j:K_1\times\cdots\times K_n\to Y_j$ are continuous operators, then for any $\epsilon>0$:
\begin{enumerate}[(i)]
    \item There exist positive integers $p_i$, $q_i$, continuous vector functions $\mathbf{g}_i\in C(\R^{q_i},\R^{p_i})$, and $\mathbf{u}_j\in Y_j^{p_1\times p_2\times\cdots\times p_n}$, such that
\begin{equation*}
    \sup_{v_i\in K_i}\norm{\mathcal{G}_j(v_1,\cdots,v_n)-\mathbf{u}_j\langle\mathbf{g}_1(\varphi_{q_1}^1(v_1)),\cdots,\mathbf{g}_n(\varphi_{q_n}^n(v_n))\rangle}<\epsilon,\quad j=1,\cdots,m.
\end{equation*}
    \item There exist positive integers $p_i,q_i,r_j$, continuous vector functions $\mathbf{g}_i\in C(\R^{q_i},\R^{p_i})$, $\mathbf{u}_j\in Y_j^{r_j}$, and $W_j\in\R^{p_1\times\cdots\times p_n\times r_j}$, such that
    \begin{equation*}
        \sup_{v_i\in K_i}\norm{\mathcal{G}_j(v_1,\cdots,v_n)-W_j\langle\mathbf{g}_1(\varphi_{q_1}^1(v_1)),\cdots,\mathbf{g}_n(\varphi_{q_n}^n(v_n)),\mathbf{u}_j\rangle}<\epsilon,\quad j=1,\cdots,m.
    \end{equation*}
    If $\{e_k^j\}$ is a Schauder basis for $Y_j$, we can further have $\mathbf{u}_j=(e_1^j,e_2^j,\cdots,e_{r_j}^j)^T$.
    \item There exist positive integers $p,q_i$, continuous functions $g_{k}^i\in C(\R^{q_{i}})$, $u_k^j\in Y_j$, such that
    \begin{equation*}
    \sup_{v_i\in K_i}\norm{\mathcal{G}_j(v_1,\cdots,v_n)-\sum_{k=1}^{p}g_k^1(\varphi_{q_1}^1(v_1))\cdots g_k^n(\varphi_{q_n}^n(v_n))\cdot u_k^j}<\epsilon,\quad j=1,\cdots,m.
\end{equation*}
\end{enumerate}
\end{corollary}

Corollary \ref{cor:multiple_outputs} can be easily obtained by replacing $Y$ with $Y_1\times\cdots\times Y_m$ in Theorem \ref{thm:approximation}. We list it here to emphasize that multiple operators defined on the same product spaces can share the same $\mathbf{g}_i$ or $g_j^i$, which indicates a practical approximation method for operators mapping from $X_1\times\cdots\times X_n$ to $Y_1\times\cdots\times Y_m$.

\begin{corollary}[Linear case] \label{cor:linear}
In Theorem \ref{thm:approximation} and Corollary \ref{cor:equiform}, if $\mathcal{G}$ is linear with respect to $v_i$, then linear $\mathbf{g}_i$ and $g_j^i$ is sufficient.
\end{corollary}

By Corollary \ref{cor:linear}, if we know the operator is linear with respect to $v_i$, then we can choose $\mathbf{g}_i$ or $g_j^i$ as linear maps in practice to make the learning procedure easier and generalize better.

\begin{property} \label{pro:rm_lin}
In Theorem \ref{thm:approximation} and Corollary \ref{cor:equiform}, if $\mathbf{g}_i=W_i\cdot\mathbf{h}_i$ for $W_i\in\R^{p_i\times h_i}$ and $\mathbf{h}_i\in C(\R^{q_i},\R^{h_i})$, then Eqs.~(\ref{eq:approximation}) and (\ref{eq:symm}) can be rewritten as
\begin{enumerate}[(i)]
    \item $\sup_{v_i\in K_i}\norm{\mathcal{G}(v_1,\cdots,v_n)-\tilde{\mathbf{u}}\langle\mathbf{h}_1(\varphi_{q_1}^1(v_1)),\cdots,\mathbf{h}_n(\varphi_{q_n}^n(v_n))\rangle}<\epsilon,$
    \item $\sup_{v_i\in K_i}\norm{\mathcal{G}(v_1,\cdots,v_n)-\tilde{W}\langle\mathbf{h}_1(\varphi_{q_1}^1(v_1)),\cdots,\mathbf{h}_n(\varphi_{q_n}^n(v_n)),\mathbf{u}\rangle}<\epsilon,$
\end{enumerate}
respectively, for a new $\tilde{\mathbf{u}}\in Y^{h_1\times\cdots\times h_n}$ and a new $\tilde{W}\in\R^{h_1\times\cdots\times h_n\times r}$.
\end{property}

Property \ref{pro:rm_lin} shows that the linear output layers of $\mathbf{g}_{i}$ are allowed to be removed without loss of universality. For example, when $\mathbf{g}_i$ are chosen as FNNs, then we can eliminate the redundant parameters of the last linear layer.

\begin{corollary}[Universal approximation theorem for functions] \label{cor:function}
Assume that $f:K_1\times\cdots\times K_n\to\R$ is a continuous function for compact $K_i\subset \R^{q_i}$, $\sigma$ is an activation function which satisfies the requirements for approximation theorem of fully-connected neural networks, then for any $\epsilon>0$:
\begin{enumerate}[(i)]
\item There exist integers $p_i$, weights $W_i\in\R^{p_i\times q_i}$, $W\in\R^{p_1\times\cdots\times p_n}$ and biases $b_i\in\R^{p_i}$, such that
\begin{equation*}
    \norm{f-W\langle \sigma(W_1(\cdot)+b_1),\cdots,\sigma(W_n(\cdot)+b_n)\rangle}_{C(K_1\times\cdots\times K_n)}<\epsilon.
\end{equation*}
\item There exist integer $p$, weights $W_i\in\R^{p_i\times q_i}$, $w_j^i\in\R^{1\times p_i}$ and biases $b_i\in\R^{p_i}$, such that
\begin{equation*}
    \norm{f-\sum_{j=1}^p \left( w_j^1\sigma(W_1(\cdot)+b_1)\right) \cdot \left( w_j^2\sigma(W_2(\cdot)+b_2)\right) \cdots \left( w_j^n\sigma(W_n(\cdot)+b_n)\right)}_{C(K_1\times\cdots\times K_n)}<\epsilon.
\end{equation*}
\end{enumerate}
\end{corollary}

When $n=1$, Corollary \ref{cor:function} degenerates to the classical universal approximation theorem for fully-connected neural networks with one hidden layer. Generally speaking, $V_1\times\cdots\times V_n$ can be regarded as a compact $V\in\R^{q_1+\cdots+q_n}$, and thus $f$ can also be approximated by FNNs. Compared to FNNs, the two new architectures in Corollary \ref{cor:function} divide the input components into several different groups.

\subsection{View through the tensor product of Banach spaces}
\label{sec:tensor}

We have presented all the main theorems and related analysis. Here, we provide another view of the theory through the tensor product of Banach spaces, which is optional to the reader, and it will not hinder the understanding of the subsequent content without reading this section. We also refer the reader to \cite{ryan2002introduction} for more details.

Recall that we aim to learn a continuous operator
\begin{equation*}
    \mathcal{G} \in C(K, Y),\quad K=K_1\times\cdots\times K_n,
\end{equation*}
where $K_i$ is a compact set in a Banach space $X_i$. By the injective tensor product, we have
\begin{equation*}
    C(K,Y)\cong C(K)\itp Y,
\end{equation*}
with the canonical linear map defined as
\begin{equation*}
\begin{split}
    J:C(K)\otimes Y & \to C(K,Y) \\
    \sum_{j=1}^p f_j\otimes u_j & \mapsto\sum_{j=1}^p f_j\cdot u_j
\end{split}
\end{equation*}
for representation $\mu=\sum_{j=1}^p f_j\otimes u_j\in C(K)\otimes Y$. Here, $C(K)\itp Y$ is the completion of $C(K)\otimes Y$ with the injective norm $\varepsilon(\mu)=\sup_{v\in K}\norm{\sum_{j=1}^pf_j(v)u_j}$, and we have the isometric isomorphism between $C(K)\itp Y$ and $C(K,Y)$, for convenience, still denoted as $J$. Then for $\mathcal{G}\in C(K,Y)$ and any $\epsilon>0$, there exists $\sum_{j=1}^p f_j\otimes u_j$ such that
\begin{equation}\label{eq:isometric}
    \norm{\mathcal{G}-\sum_{j=1}^p f_j\cdot u_j}_{C(K,Y)}=\varepsilon\left(J^{-1}\mathcal{G}-\sum_{j=1}^p f_j\otimes u_j\right)<\epsilon.
\end{equation}
Furthermore, by repeating the following decomposition
\begin{equation*}
\begin{split}
    C(K_1\times\cdots\times K_n)
    \cong&C(K_1\times\cdots\times K_{n-1},C(K_n)) \\
    \cong&C(K_1\times\cdots\times K_{n-1})\itp C(K_n), \\
\end{split}
\end{equation*}
we obtain
\begin{equation}\label{eq:key}
    C(K_1\times K_2\times\cdots\times K_n,Y)\cong C(K_1)\itp C(K_2)\itp\cdots\itp C(K_n)\itp Y.
\end{equation}
Similar to Eq.~(\ref{eq:isometric}), we have
\begin{equation}\label{eq:ele2}
\norm{\mathcal{G}-\sum_{j=1}^p f_{j}^1\cdot f_{j}^2\cdots f_{j}^n\cdot u_j}_{C(K,Y)}<\epsilon
\end{equation}
for some $f_j^i\in C(K_i)$ and $u_j\in Y$. Note that $C(K_i)$, as a continuous function space on compact $K_i$, has a Schauder basis, denoted as $\{g_k^i\}_{k=1}^{\infty}$. Let $f_j^i=\sum_{k=1}^\infty\alpha_{jk}^{i}g_k^i$, then there exist positive integers $p_i$ such that
\begin{equation}\label{eq:app2}
\norm{\mathcal{G}-\mathbf{u}\langle\mathbf{g}_1,\cdots,\mathbf{g}_n\rangle}_{C(K,Y)}<\epsilon    
\end{equation}
for $\mathbf{g}_i = \left(g_1^i,\cdots,g_{p_i}^i\right)^T$, $\mathbf{u} = \left(\sum_{j=1}^p\alpha_{jk_1}^1\cdots\alpha_{jk_n}^nu_j\right)_{p_1\times\cdots\times p_n}$. Furthermore, if $Y$ has a Schauder basis $\{e_k\}$ and $u_j=\sum_{k=1}^\infty\beta_{jk}e_k$, there exists a $r$ such that
\begin{equation}\label{eq:sym2}
\norm{\mathcal{G}-W\langle\mathbf{g}_1,\cdots,\mathbf{g}_n,\mathbf{e}\rangle}_{C(K,Y)}<\epsilon
\end{equation}
for $\mathbf{e}=(e_1,\cdots,e_r)^T$, $W = \left(\sum_{j=1}^p\alpha_{jk_1}^1\cdots\alpha_{jk_n}^n\beta_{jk_{n+1}}\right) \in \R^{p_1\times\cdots\times p_n\times r}$. Hence, we have also obtained the three approximation formulas (\ref{eq:ele2}--\ref{eq:sym2}) corresponding to Eqs.~(\ref{eq:approximation}--\ref{eq:elementwise}), with additional information that the components of $\mathbf{g}_i$ and $\mathbf{e}$ are all basis functions.

Next we analyze the complexity of the approximation in terms of tensor rank. The operator $\mathcal{G}$ in Eq.~(\ref{eq:sym2}) is discretely represented by a tensor $W\in\R^{p_1\times\cdots\times p_n\times r}$, since $\mathbf{g}_i$ and $\mathbf{e}$ are basis. The number of parameters of $W$ grows exponentially with respect to $n$, so directly using $W$ in computation is too expensive for large $n$. However, if we rewrite $W$ as
\begin{equation*}
    W = \left(\sum_{j=1}^p\alpha_{jk_1}^1\cdots\alpha_{jk_n}^n\beta_{jk_{n+1}}\right) = \sum_{j=1}^p\mathbf{a}_j^1\otimes\cdots\otimes\mathbf{a}_j^n\otimes\mathbf{b}_j
\end{equation*}
for $\mathbf{a}_j^i=(\alpha_{jk}^i)\in\R^{p_i}$, $\mathbf{b}_j=(\beta_{jk})\in\R^r$, then we have
\begin{equation} \label{eq:W_tensor}
    W\langle\mathbf{g}_1,\cdots,\mathbf{g}_n,\mathbf{e}\rangle = \sum_{j=1}^pf_{j}^1\cdot f_j^2\cdots f_{j}^n\cdot u_j
\end{equation}
for $f_j^i=\sum_{k=1}^{p_i}\alpha_{jk}^ig_{k}^i$, $u_j=\sum_{k=1}^r\beta_{jk}e_k$. Here, $W$ in Eq.~\eqref{eq:W_tensor} is a tensor of rank at most $p$. From this point of view, Eq.~(\ref{eq:ele2}) also gives a low-rank approximation by the tensor
$$\mu=\sum_{j=1}^p f_{j}^1\otimes\cdots \otimes f_{j}^n\otimes u_j$$
of rank at most $p$. We note that it is usually difficult to determine the rank of high-order tensors, which is NP-hard \cite{johan1990tensor}, but in some cases there exist some relationships between the dimension of $W$ and its rank $p$. For example, if $W\in\R^{p_1\times p_2\times p_3}$, then the rank of $W$ has an upper bound \cite{kolda2009tensor,kruskal1989rank}:
\begin{equation*}
    {\rm rank}(W)\leq\min\{p_1p_2,p_1p_3,p_2p_3\}.
\end{equation*}


In short, we have a more general viewpoint for our results. Assume that $Y=C(K_0)$ for a compact set $K_0$ in a Banach space $X_0$, then depending on the level of decomposition applied, we have the following cases:
\begin{numcases}
{C(K_1\times K_2\times\cdots\times K_n,Y)=}
C(K_1\times K_2\times\cdots\times K_n\times K_0), & (standard NN) \label{eq:fnn} \\
C(K_1\times K_2\times\cdots\times K_n)\itp C(K_0), & (DeepONet) \label{eq:deeponet} \\
C(K_1)\itp C(K_2)\itp\cdots\itp C(K_n)\itp C(K_0). & (MIONet) \label{eq:MIONet}
\end{numcases}
We discuss how the three different representations lead to different network architectures as follows.
\begin{itemize}
    \item In Eq.~\eqref{eq:fnn}, we first combine together all the inputs from all the spaces and then pass them into a function (i.e., a machine learning model) to approximate $\mathcal{G}$. When we restrict the model to be a neural network, then it is a standard NN such as FNN, residual neural network (ResNet), convolutional neural network (CNN), etc.
    \item In Eq.~\eqref{eq:deeponet}, we first split the input and output space, and have one model for the input space and one model for the output space, and then combine them. When both models are standard NNs, this leads to the same architecture as DeepONet \cite{lu2021learning}, but DeepONet only considers $n=1$.
    \item In Eq.~\eqref{eq:MIONet}, we split all the spaces with one model for each one, and then combine them to compute the output. This leads to our proposed MIONet in Section \ref{sec:method}, where each model is a standard NN.
\end{itemize}


\section{Operator regression methods}
\label{sec:method}

Based on our theory, we propose a new neural operator, MIONet, for learning multiple-input operator regression.

\subsection{Network architectures}
\label{sec:MIONet}

The architectures of MIONet are designed based on Theorem \ref{thm:approximation} and Corollary \ref{cor:equiform} with $Y=C(K_0)$ for a compact set $K_0\subset\R^d$. We design two slightly different versions of MIONet according to different formulas as follows.

\paragraph{MIONet (high-rank).}
We first construct the architecture according to Eqs.~(\ref{eq:approximation}) and (\ref{eq:symm}). Note that the architecture induced by Eq.~(\ref{eq:symm}) is technically equivalent to Eq.~(\ref{eq:approximation}) as we discussed in Section \ref{sec:nonlinear_operators}. Specifically, we use $\mathbf{f}\in C(K_0,\R^{p_1\times\cdots\times p_n})$ to denote the $\mathbf{u}$ in Eq.~(\ref{eq:approximation}), and we approximate $\mathbf{g}_i$ and $\mathbf{f}$ by independent neural networks denoted by $\tilde{\mathbf{g}}_i$ (called branch net $i$) and $\tilde{\mathbf{f}}$ (called trunk net). We also add a trainable bias $b\in\R$ according to Corollary \ref{cor:bias}. Then the network is
\begin{equation} \label{eq:MIONet_high}
    \tilde{\mathcal{G}}(v_1,\cdots,v_n)(y) = \underbrace{\tilde{\mathbf{f}}(y)}_{\text{trunk}} \left\langle \underbrace{\tilde{\mathbf{g}}_1(\varphi_{q_1}^1(v_1))}_{\text{branch}_1}, \cdots, \underbrace{\tilde{\mathbf{g}}_n(\varphi_{q_n}^n(v_n))}_{\text{branch}_n} \right\rangle + b.
\end{equation}
MIONet has $n$ independent branch nets and one trunk net. The $i$th branch net $\tilde{\mathbf{g}}_i$ encodes the input function $v_i$, and the trunk net $\tilde{\mathbf{f}}$ encodes the input $y$. The output tensor of the trunk net has a high rank as we discussed in Section \ref{sec:tensor}. We note that the last linear layer of each branch net can be removed by Property \ref{pro:rm_lin} to reduce the number of parameters. As a special case, if the image space $Y$ is finite dimensional, we also show the corresponding MIONet in Appendix \ref{app:finite_im}.

\begin{figure}[htbp]
    \centering
    \includegraphics[width=0.9\textwidth]{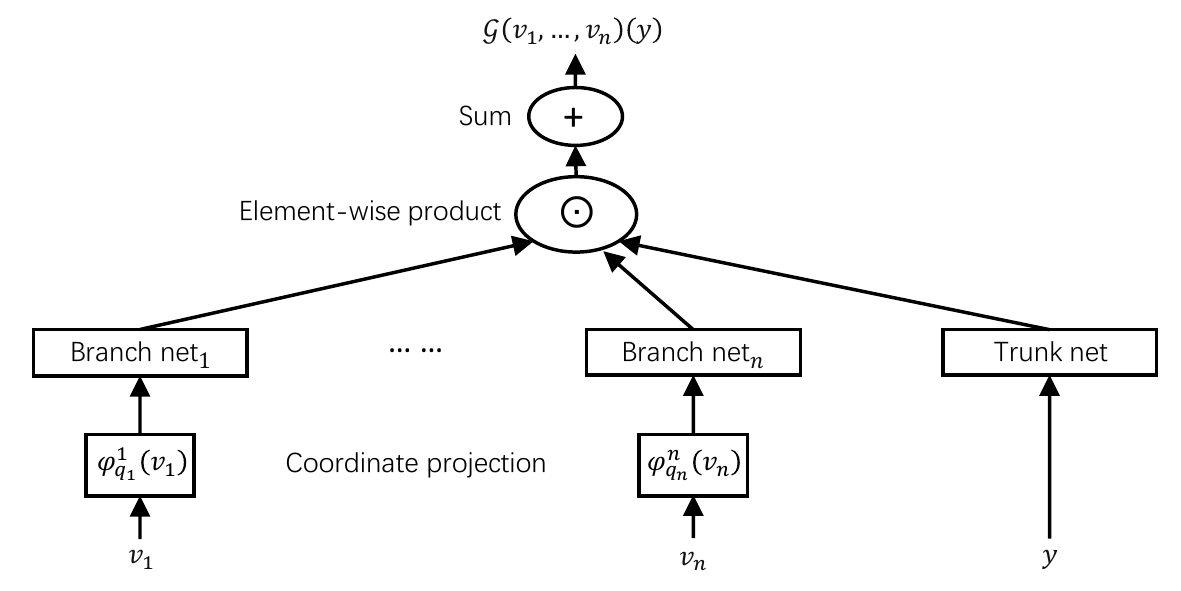}
    \caption{\textbf{Architecture of MIONet.} All the branch nets and the trunk net have the same number of outputs, which are merged together via the Hadamard product and then a summation.}
    \label{fig:architecture}
\end{figure}

\paragraph{MIONet (low-rank; the default version).}
We then construct MIONet according to Eq.~(\ref{eq:elementwise}). Specifically, $\mathbf{g}_i=(g_1^i,\cdots,g_p^i)^T\in C(\R^{q_i},\R^p)$ and $\mathbf{f}=(u_1,\cdots,u_p)^T\in C(K_0,\R^p)$ are approximated by neural networks $\tilde{\mathbf{g}}_i$ (called branch net $i$) and $\tilde{\mathbf{f}}$ (called trunk net). Then the network (Fig.~\ref{fig:architecture}) is
\begin{equation} \label{eq:MIONet_low}
    \tilde{\mathcal{G}}(v_1,\cdots,v_n)(y) = \mathcal{S}\left( \underbrace{\tilde{\mathbf{g}}_1(\varphi_{q_1}^1(v_1))}_{\text{branch}_1} \odot\cdots\odot \underbrace{\tilde{\mathbf{g}}_n(\varphi_{q_n}^n(v_n))}_{\text{branch}_n} \odot \underbrace{\tilde{\mathbf{f}}(y)}_{\text{trunk}} \right) + b,
\end{equation}
where $\odot$ is the Hadamard product (i.e., element-wise product), $\mathcal{S}$ is the summation of all the components of a vector, and $b\in\R$ is a trainable bias. This MIONet is a low-rank version of the MIONet (high-rank) above, which greatly reduces the number of parameters of the trunk net. Furthermore, if the output function is also defined on a product space, e.g., $C(K_0)=C(K_{01}\times K_{02})$, we can choose to further decompose it into $C(K_{01})\itp C(K_{02})$, and the corresponding MIONet can be built similarly.

\paragraph{Connections to DeepONet.}
Our proposed MIONet is related to DeepONet. When there is only one input function, i.e., $n=1$, MIONet becomes DeepONet with one branch net and one trunk net.

\paragraph{Remark.}
In this work, we mainly consider MIONet (low-rank), as MIONet (high-rank) is computationally expensive. We note that MIONet is a high-level architecture, where the neural networks $\tilde{\mathbf{g}}_i$ and $\tilde{\mathbf{f}}$ can be chosen as any valid NNs, such as FNN, ResNet, CNN, etc., depending on the specific problem. All the techniques developed for DeepONet in \cite{lu2021comprehensive} can be directly used for MIONet. For example, we can encode the periodicity in the trunk net to ensure that the predict functions from MIONet are always periodic. We refer the reader to \cite{lu2021comprehensive} for more details.

\subsection{Other computational details}

\paragraph{Data.}
One data point in the dataset is comprised of input functions and their corresponding output function, i.e., $(v_1,\cdots,v_n,\mathcal{G}(v_1,\cdots,v_n))$. In the first step of MIONet, we project the input functions $v_i$ onto finite-dimensional spaces as stated in Eqs.~\ref{eq:MIONet_high} and \ref{eq:MIONet_low}, which can be done separately before the network training. Hence, the network input in practice is $(\varphi_{q_1}^1(v_1),\cdots,\varphi_{q_n}^n(v_n))$, and then the dataset takes the form
\begin{equation}\label{eq:data}
    \mathcal{T} = \left\{\left( \underbrace{\varphi_{q_1}^1(v_1^k)}_{\text{branch}_1}, \cdots, \underbrace{\varphi_{q_n}^n(v_n^k)}_{\text{branch}_n}, \underbrace{y_k}_{\text{trunk}}, \underbrace{\mathcal{G}\left(v^k_1,\cdots,v^k_n\right)\left(y_k\right)}_{\text{output}} \right) \right\}_{k=1}^N,
\end{equation}
where $y_k\in K_0\subset\R^d$ is a single point location in the domain of the output function.

\paragraph{Training.}
For a training dataset $\mathcal{T}$, in this study we use a standard mean squared error (MSE):
\begin{equation*}
    \text{MSE} = \frac{1}{N}\sum_{k=1}^{N}|s_k-\tilde{\mathcal{G}}(v_1^k,\cdots,v_n^k)(y_k)|^2.
\end{equation*}
We also provides alternative losses via numerical integration in Appendix \ref{app:num_int}.

\paragraph{Inference.}
For new input functions $v_1,\cdots,v_n$, the prediction is simply given by $\tilde{\mathcal{G}}(v_1,\cdots,v_n)$. We note that $\tilde{\mathcal{G}}(v_1,\cdots,v_n)$ is a function given by neural networks, which can be evaluated at arbitrary points without interpolation.

\section{Numerical results}
\label{sec:results}

To demonstrate the capability of MIONet, we learn three different operators of ODEs and PDEs. In the experiments, we directly evaluate the function values at uniform grid points as the input of branch nets, i.e., each $\varphi$ takes 100 equidistant sampling points in $[0,1]$ for each input function. The branch and trunk nets are all chosen as fully-connected neural networks (FNNs) unless noted otherwise. Each branch or trunk net has the same number of neurons (i.e., width) for each layer. The activation in all networks is set to ReLU. We train all the networks by the Adam optimizer \cite{adam2015}. To evaluate the performance of the networks, we compute the $L^2$ relative error of the predictions, and for each case, five independent training trials are performed to compute the mean error and the standard deviation. The code in this study is implemented by using the library DeepXDE \cite{lu2021deepxde}, and is publicly available from the GitHub repository \url{https://github.com/lu-group/mionet}.

MIONet is the first neuron operator designed for multiple inputs with theoretical guarantees, so there is no other network that we can directly compare MIONet with. In order to compare MIONet with DeepONet, we simply concatenate all the input functions together as the input of DeepONet branch net.

\subsection{An ODE system}

We first consider a nonlinear ODE system:
\begin{equation*}
\begin{split}
    \frac{du_{1}}{dt} = u_{2},\quad \frac{du_{2}}{dt} = -f_{1}(t)\sin(u_{1}) + f_{2}(t),\quad t\in [0,1], 
\end{split}
\end{equation*}
with an initial condition $u_1(0)=u_2(0)=0$. We learn the operator mapping from $f_{1}$ and $f_{2}$ to one of the ODE solutions $u_{1}$:
$$\mathcal{G}: (f_1, f_2) \mapsto u_1.$$

To generate the dataset, $f_1$ and $f_2$ are both sampled from a Gaussian random field (GRF) 
$$\mathcal{GP}(0, k_l(x_1, x_2)),$$
where the covariance kernel $k_l(x_1, x_2) = \exp(-\|x_1-x_2\|^2/2l^2)$ is the Gaussian kernel with a length-scale parameter $l$. Here, we choose $l=0.2$. We set the number of functions in the training/test set to 1000, and for each couple of $(f_1,f_2)$ we get the numerical solution of $u_1$ at 100 equidistant grid points in $[0, 1]$. We train the networks for 100000 epochs with learning rate 0.001.

MIONet has the $L^2$ relative error of 1.69\% (Table \ref{tab:ode}), which outperforms DeepONet with almost the same number of parameters (2.41\%). We also perform a grid search for the depth and width to find the best accuracy of DeepONet, and the best DeepONet accuracy is 2.26\%, which is still worse than MIONet.

\begin{table}[htbp]
  \centering
    \begin{tabular}{c|c c c c}
    \toprule
     & Depth & Width & No. of parameters & $L^2$ relative error \\
    \midrule
    MIONet  & 2  &  200 & 161K  & $1.69 \pm 0.13\%$ \\
    DeepONet (same size) & 2 & 312 & 161K & $2.41 \pm 0.27\%$ \\
    DeepONet (best) & 2 & 300 & 151K & $2.26 \pm 0.14\%$ \\
    \bottomrule
    \end{tabular}
    \caption{\textbf{MIONet and DeepONet for an ODE system.} DeepONet (same size) has the same number of parameters as MIONet. DeepONet (best) is the best result chosen from depth 2--5 and width 100--400.}
  \label{tab:ode}
\end{table}

\subsection{A diffusion-reaction system}

We consider a nonlinear diffusion-reaction system
\begin{equation*}
    \frac{\partial u}{\partial t} = \frac{\partial }{\partial x}\left(D(x)\frac{\partial u}{\partial x}\right) + ku^2+g(x),\quad x \in [0,1],\ t\in [0,1], 
\end{equation*}
with zero initial and boundary conditions, where $D(x)=0.01(|f(x)|+1)$ and $k = 0.01$ is the reaction rate. We learn the operator
$$\mathcal{G}: (D, g) \mapsto u.$$

In the dataset, $f$ and $g$ are generated by GRF with length scale 0.2. We set the number of couples of $(D,g)$ in the training dataset to 1000 and test dataset to 5000, and for each couple we solve $u$ in a grid with $100\times 100$ equidistant points. We train each case for 100000 epochs with learning rate 0.001.

\begin{figure}[htbp]
    \centering
    \includegraphics[width=0.8\textwidth]{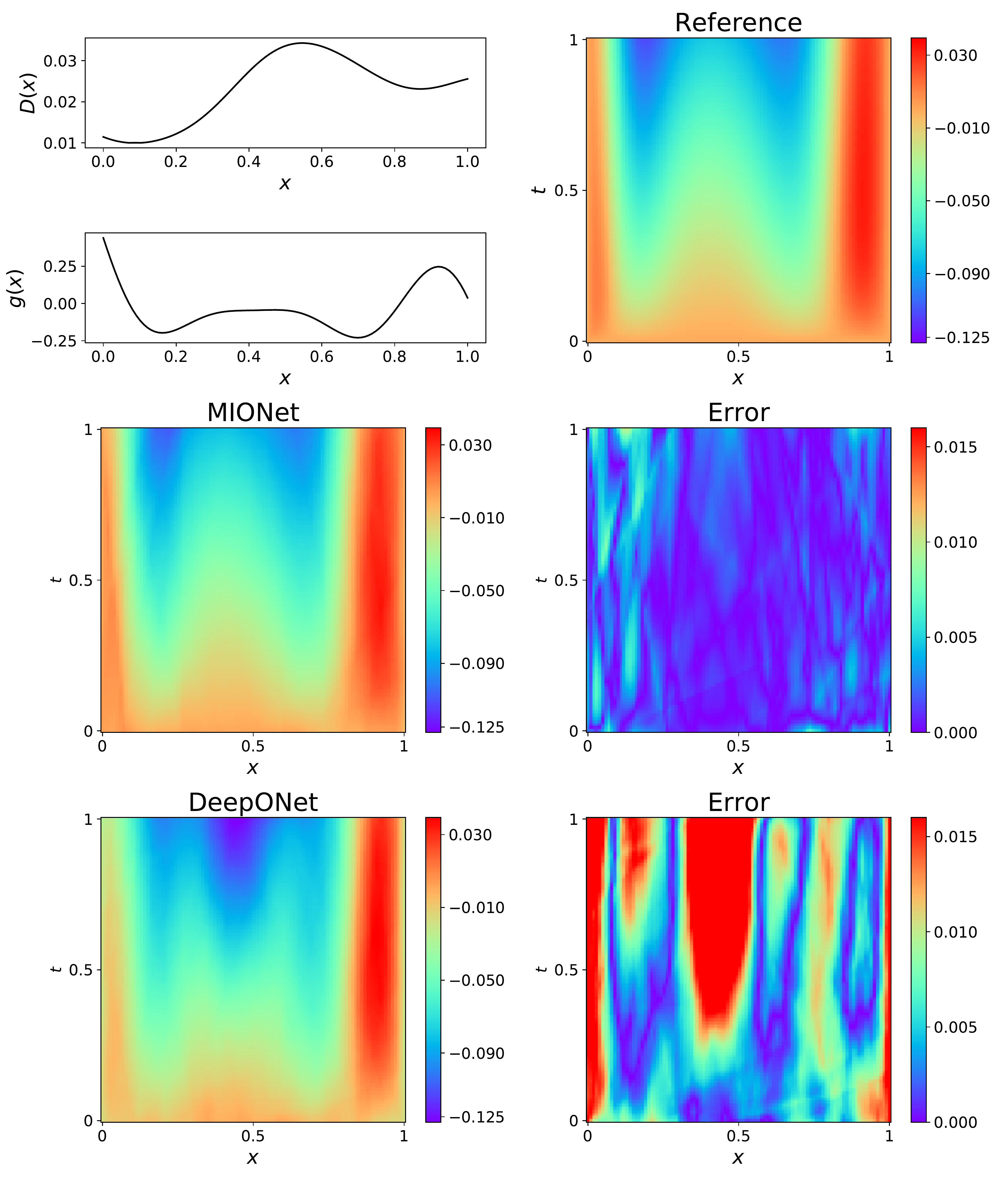}
    \caption{\textbf{Example of the diffusion-reaction system.} (Top) Examples of the input functions (left) and the reference solution (right). (Middle) MIONet prediction and corresponding absolute error. (Bottom) DeepONet prediction and corresponding absolute error.}
    \label{fig:dr}
\end{figure}

The error of MIONet is significantly less than that of DeepONet of similar size and also the best DeepONet (Table \ref{tab:dr}). In Fig.~\ref{fig:dr}, we show an example of the inputs and the corresponding PDE solution. We also show the prediction and point-wise error of DeepONet and MIONet.

\begin{table}[htbp]
  \centering
    \begin{tabular}{c|c c c c c}
    \toprule
     & Depth & Width & Parameters & $L^2$ relative error \\
    \midrule
    MIONet & 2  & 200  & 161K  & $ 1.97 \pm 0.11 \%$ \\
    DeepONet (same size) & 2 & 312 & 161K & $5.25 \pm 0.38\%$ \\
    DeepONet (best) & 2 & 400 & 242K & $5.18 \pm 0.11\%$ \\
    \bottomrule
    \end{tabular}%
    \caption{\textbf{MIONet and DeepONet for a diffusion-reaction system.} DeepONet (same size) has the same number of parameters as MIONet. DeepONet (best) is the best DeepONet result chosen from depth 2--5 and width 100--400.}
  \label{tab:dr}%
\end{table}%

\subsection{An advection-diffusion system}
\label{sec:advd}

We consider an advection-diffusion system
\begin{equation*}
    \frac{\partial u}{\partial t} + \frac{\partial u}{\partial x} - D(x) \frac{\partial^2 u}{\partial x^2} = 0,\quad  x \in [0,1],\ t\in [0,1], 
\end{equation*}
with the periodic boundary condition and initial condition $u_0(x)=u(x,0) = f_1(\sin^2(\pi x))$, where $D(x)= 0.05|f_2(\sin^2(\pi x)|+0.05$ is the diffusion coefficient. We aim to learn the operator
$$\mathcal{G}: (D, u_0) \mapsto u.$$

In the dataset, $f_1$ and $f_2$ are sampled from a GRF with the length scale 0.5. The training/test dataset consists of 1000 couples of $(D,u_0)$. For each $(D,u_0)$, we solve the solution $u$ numerically in a grid of size $100 \times 100$, and randomly select 100 values of $u$ out of the 10000 grid points. We train each case for 100000 epochs with learning rate 0.0002.

\begin{figure}[htbp]
    \centering
    \includegraphics[width=0.85\textwidth]{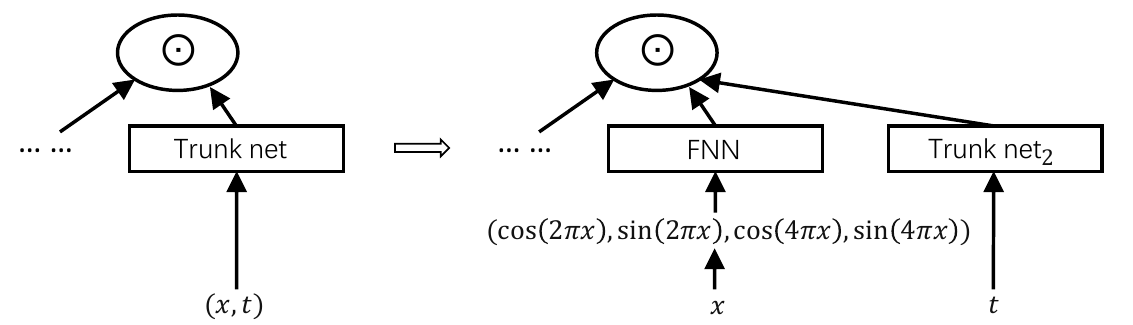}
    \caption{\textbf{Architecture of the modified MIONet for the advection-diffusion system.} There are two trunk nets, one for $x$ and one for $t$. The trunk net of $x$ has a periodic layer.}
    \label{fig:architecture_periodic}
\end{figure}

Here, we show how to encode the prior information of this problem. Since the operator $\mathcal{G}$ is linear with respect to the initial condition $u_0$, we choose the branch net for $u_0$ in MIONet to be a linear network, i.e., a linear layer without bias. Moreover, because the solution $u$ is periodic with respect to $x$, we decompose the single trunk net to two independent networks, one for $x$ and one for $t$. For the trunk net of $x$, we apply a periodic layer as the input of FNN \cite{lu2021comprehensive}:
\begin{equation*}
    {\rm Trunk}(x)={\rm FNN}(\cos(2\pi x), \sin(2\pi x),\cos(4\pi x), \sin(4\pi x)),\quad x\in\R.
\end{equation*}
It is easy to check that by using these $\cos$ and $\sin$ features, MIONet is automatically periodic with respect to $x$. We present the illustration of the modified MIONet in Fig.~\ref{fig:architecture_periodic}.

\begin{figure}[htbp]
    \centering
    \includegraphics[width=1.0\textwidth]{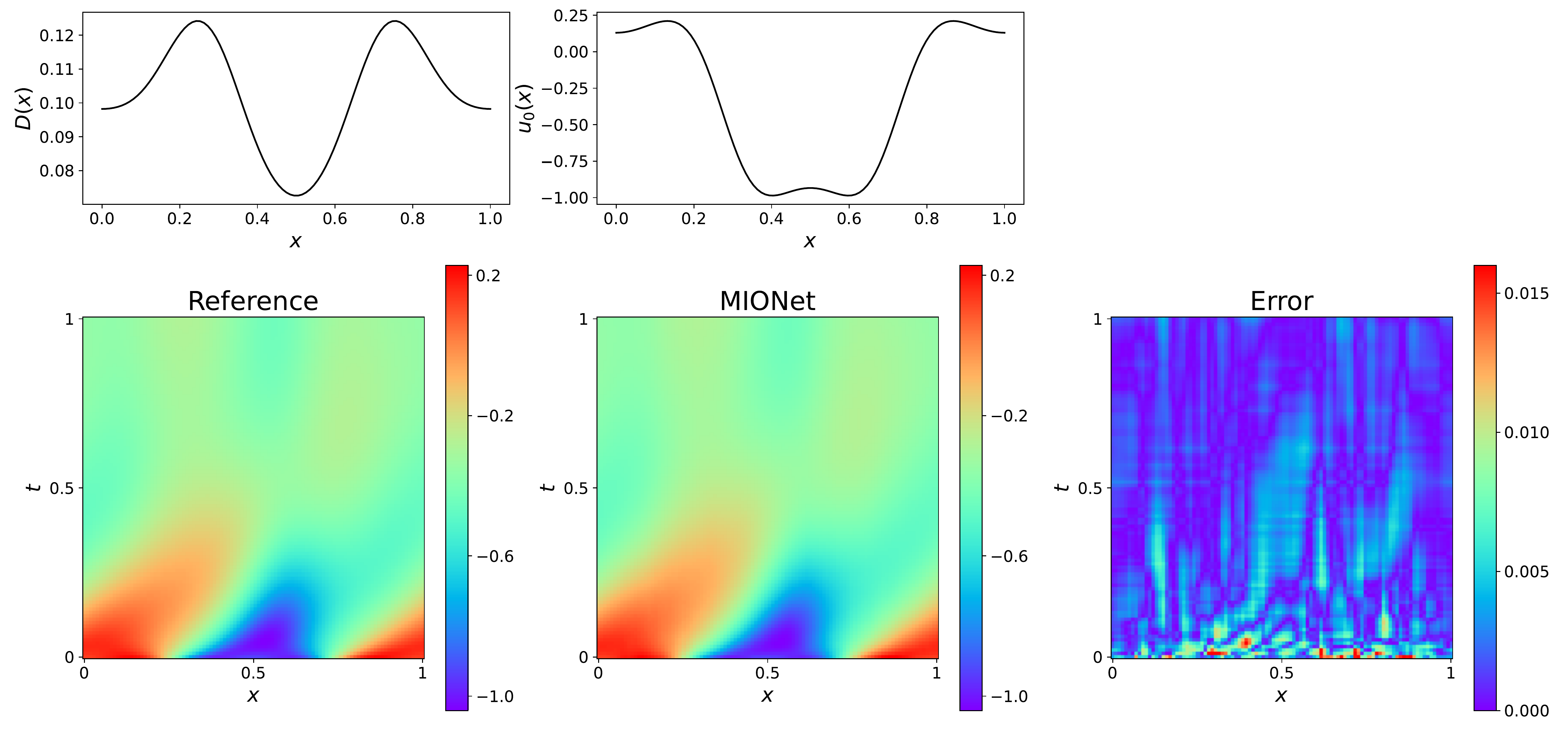}
    \caption{\textbf{Prediction of MIONet (periodic) for the advection-diffusion system.}}
    \label{fig:advd}
\end{figure}

The accuracy of different networks are shown in Table \ref{tab:advd}. MIONet performs significantly better than DeepONet (same size) and DeepONet (best). By encoding the periodicity information, MIONet (periodic) obtains the smallest prediction error. An example of prediction of MIONet (periodic) is shown in Fig.~\ref{fig:advd}.

\begin{table}[htbp]
  \centering
    \begin{tabular}{c|c c c c c}
    \toprule
     & Depth & Width & Parameters & $L^2$ relative error \\
    \midrule
    MIONet  & 3  & 300 & 422K & $ 1.98 \pm 0.07 \%$ \\
    MIONet (periodic)  & 3   & 248 & 422K & $ 1.29 \pm 0.09 \%$ \\
    DeepONet (same size) & 3 & 343 & 424K & $ 7.83 \pm 0.49 \%$ \\
    DeepONet (best) & 3 & 300 & 332K & $ 7.70 \pm 0.69 \% $\\
    \bottomrule
    \end{tabular}
\caption{\textbf{MIONet and DeepONet for the advection-diffusion system.} MIONet (periodic) has a periodic layer for the trunk net of $x$. MIONet, MIONet (periodic), and DeepONet (same size) have the same number of parameters. DeepONet (best) is the best DeepONet chosen from depth 2--5 and width 100--400.}
  \label{tab:advd}
\end{table}

\section{Conclusions}
\label{sec:conclusions}

In this study, we aim to learn an operator mapping from a product of multiple Banach spaces to another Banach space. Our main contribution is that for the first time, we provide universal approximation theorems for multiple-input operator regression based on tensor product of Banach spaces. Based on the theory and a low-rank tensor approximation, we propose a new network architecture, MIONet, which consists of multiple branch nets for encoding the input functions and one trunk net for encoding the domain of the output function. To show the effectiveness of MIONet, we have performed three experiments including an ODE system, a diffusion-reaction system and an advection-diffusion system. We also show that it is flexible to customize MIONet to encode the prior knowledge.

In future work, more experiments should be done to test the performance of MIONet on diverse problems. Moreover, MIONet can be viewed as an extension of DeepONet from a single branch net to multiple branch nets, and thus recent developments and extensions of DeepONet (see the discussion in the introduction) can be directly applied to MIONet. For example, similar to DeepONet with proper orthogonal decomposition (POD-DeepONet) \cite{lu2021comprehensive}, we can employ POD in MIONet to develop POD-MIONet. We can also embed physics into the loss function \cite{wang2021learning,goswami2022physics} of MIONet to develop physics-informed MIONet. These techniques will further improve the accuracy and efficiency of MIONet.

\appendix

\section{Proofs in this study}
\label{app:proofs}

\paragraph{Proof of Property \ref{pro:projection}.}

Since $X$ is a Banach space, $P_n$ is uniformly bounded by the basis constant $C$. For any $\epsilon>0$, we choose finite points $\{x_i\}_{i=1}^{k}\subset K$, such that the union of open balls $\cup_{i=1}^k B(x_i,\delta)$ covers $K$, where $\delta=\frac{\epsilon}{2(1+C)}$. There exists a large integer $m\in\mathbb{N}^*$ such that $\norm{x_i-P_n(x_i)}<\frac{\epsilon}{2}$ holds for all $1\leq i\leq k$ and $n\geq m$. When $n\geq m$, for any $x\in K$, assume that $x\in B(x_j,\delta)$, then
\begin{equation*}
\begin{split}
\norm{x-P_n(x)}&=\norm{(I-P_n)(x-x_j)+x_j-P_n(x_j)}\\
&\leq\norm{I-P_n}\cdot\norm{x-x_j}+\norm{x_j-P_n(x_j)}\\&<(1+C)\cdot\frac{\epsilon}{2(1+C)}+\frac{\epsilon}{2}\\&=\epsilon.
\end{split}
\end{equation*}
\qed

\paragraph{Proof of Lemma \ref{lem:functional}.}

As $\mathcal{G}$ is uniformly continuous on $K_1\times\cdots\times K_n$, there exists a $\delta>0$ such that $\|\mathcal{G}(v_1,\cdots,v_n)-\mathcal{G}(v'_1,\cdots,v'_n)\|<\epsilon$ holds for all $v_i,v'_i\in K_i$, $\|v_i-v'_i\|<\delta$, $1\leq i\leq n$. Due to the compactness of $K_i$, we can choose $\{\nu_j^i\}_{j=1}^{p_i}\subset K_i$ such that
\begin{equation} \label{eq:cover}
    \bigcup_{j=1}^{p_i}B(\nu_j^i, \delta)\supset K_i,
\end{equation}
where $B(\nu_j^i, \delta)$ denotes the open ball centered at $\nu_j^i$ with radius $\delta$, $1\leq i\leq n$. Now define $\tilde{\mathbf{g}}_i:X_i\to\R^{p_i}$ as
\begin{equation*}
    \tilde{\mathbf{g}}_i(x)=(\mbox{ReLU}(\delta-\|x-\nu_1^i\|),\mbox{ReLU}(\delta-\|x-\nu_2^i\|),\cdots,\mbox{ReLU}(\delta-\|x-\nu_{p_i}^i\|))^T,
\end{equation*}
and $\hat{\mathbf{g}}_i(x):X_i\to\R^{p_i}$ as
\begin{equation*}
    \hat{\mathbf{g}}_i(x)=\frac{\tilde{\mathbf{g}}_i(x)}{\|\tilde{\mathbf{g}}_i(x)\|_1+d(x,K_i)},
\end{equation*}
where ${\rm ReLU}(x):=\max(x,0)$, and $d(x,K_i):=\inf_{x'\in K_i}\norm{x-x'}$ represents the distance between $x$ and $K_i$. $\hat{\mathbf{g}}_i$ is in fact the normalization of $\tilde{\mathbf{g}}_i$ on $K_i$, and the condition (\ref{eq:cover}) guarantees that $\hat{\mathbf{g}}_i(x)$ is well defined everywhere, i.e., $\|\tilde{\mathbf{g}}_i(x)\|_1$ is nonzero on $K_i$ while $d(x,K_i)$ is nonzero outside $K_i$. Moreover, define
$\mathbf{u}\in Y^{p_1\times p_2\times\cdots\times p_n}$ as
\begin{equation*}
    \mathbf{u}=(\mathcal{G}(\nu_{j_1}^1,\nu_{j_2}^2,\cdots,\nu_{j_n}^n))_{p_1\times p_2\times\cdots\times p_n}.
\end{equation*}
We will show that the constructed $\hat{\mathbf{g}}_i$ and $\mathbf{u}$ are what we need. 

Denote $\hat{\mathbf{g}}_i=(\hat{\mathbf{g}}_i^1,\cdots,\hat{\mathbf{g}}_i^{p_i})^T$ and define $A_i[v]=\{j\in\{1,2,\cdots,p_i\}|\|\nu_j^i-v\|<\delta\}$ for $v\in  K_i$. Given arbitrarily $v_i\in K_i$, we have
\begin{equation*}
\begin{split}
    &\norm{\mathcal{G}(v_1,\cdots,v_n)-\mathbf{u}\langle\hat{\mathbf{g}}_1(v_1),\cdots,\hat{\mathbf{g}}_n(v_n)\rangle}\\
    =&\norm{\mathcal{G}(v_1,\cdots,v_n)-\sum_{j_1,\cdots,j_n}\mathcal{G}(\nu_{j_1}^1,\nu_{j_2}^2,\cdots,\nu_{j_n}^n)\cdot\hat{\mathbf{g}}_1^{j_1}(v_1)\cdots\hat{\mathbf{g}}_n^{j_n}(v_n)}\\
    =&\norm{\sum_{j_1,\cdots,j_n}\mathcal{G}(v_1,\cdots,v_n)\cdot\hat{\mathbf{g}}_1^{j_1}(v_1)\cdots\hat{\mathbf{g}}_n^{j_n}(v_n)-\sum_{j_1,\cdots,j_n}\mathcal{G}(\nu_{j_1}^1,\nu_{j_2}^2,\cdots,\nu_{j_n}^n)\cdot\hat{\mathbf{g}}_1^{j_1}(v_1)\cdots\hat{\mathbf{g}}_n^{j_n}(v_n)}\\
    =&\norm{\sum_{j_1,\cdots,j_n}(\mathcal{G}(v_1,\cdots,v_n)-\mathcal{G}(\nu_{j_1}^1,\nu_{j_2}^2,\cdots,\nu_{j_n}^n))\cdot\hat{\mathbf{g}}_1^{j_1}(v_1)\cdots\hat{\mathbf{g}}_n^{j_n}(v_n)}\\
    =&\norm{\sum_{\substack{j_i\in A_i[v_i]\\1\leq i\leq n}}(\mathcal{G}(v_1,\cdots,v_n)-\mathcal{G}(\nu_{j_1}^1,\nu_{j_2}^2,\cdots,\nu_{j_n}^n))\cdot\hat{\mathbf{g}}_1^{j_1}(v_1)\cdots\hat{\mathbf{g}}_n^{j_n}(v_n)}\\
    \leq&\sum_{\substack{j_i\in A_i[v_i]\\1\leq i\leq n}}\left\|\mathcal{G}(v_1,\cdots,v_n)-\mathcal{G}(\nu_{j_1}^1,\nu_{j_2}^2,\cdots,\nu_{j_n}^n)\right\|\cdot\hat{\mathbf{g}}_1^{j_1}(v_1)\cdots\hat{\mathbf{g}}_n^{j_n}(v_n)\\
    <&\sum_{\substack{j_i\in A_i[v_i]\\1\leq i\leq n}}\epsilon\cdot\hat{\mathbf{g}}_1^{j_1}(v_1)\cdots\hat{\mathbf{g}}_n^{j_n}(v_n)\\
    =&\epsilon.
\end{split}
\end{equation*}
\qed

\paragraph{Proof of Theorem \ref{thm:separate}.}

For any $\epsilon>0$, there exist $p_i,\hat{\mathbf{g}}_i,\mathbf{u}$ as defined in the proof of Lemma \ref{lem:functional}, such that 
\begin{equation*}
    \sup_{v_i\in K_i}\norm{\mathcal{G}(v_1,\cdots,v_n)-\mathbf{u}\langle\hat{\mathbf{g}}_1(v_1),\cdots,\hat{\mathbf{g}}_n(v_n)\rangle}<\epsilon.
\end{equation*}
Denote $M=\max_{v_i\in K_i}\norm{\mathcal{G}(v_1,\cdots,v_n)}$, and then for positive integers $q_i$,
\begin{equation*}
\begin{split}
    &\norm{\mathbf{u}\langle\hat{\mathbf{g}}_1\circ P_{q_1}^1(v_1),\cdots,\hat{\mathbf{g}}_n\circ P_{q_n}^n(v_n)\rangle-\mathbf{u}\langle\hat{\mathbf{g}}_1(v_1),\cdots,\hat{\mathbf{g}}_n(v_n)\rangle}\\
    =&\norm{\sum_{i_1,\cdots,i_n}\mathcal{G}(\nu_{i_1}^1,\nu_{i_2}^2,\cdots,\nu_{i_n}^n)\cdot(\hat{\mathbf{g}}_1^{i_1}\circ P_{q_1}^1(v_1)\cdots\hat{\mathbf{g}}_n^{i_n}\circ P_{q_n}^n(v_n)-\hat{\mathbf{g}}_1^{i_1}(v_1)\cdots\hat{\mathbf{g}}_n^{i_n}(v_n))}\\
    \leq&M\sum_{i_1,\cdots,i_n}\left|\hat{\mathbf{g}}_1^{i_1}\circ P_{q_1}^1(v_1)\cdots\hat{\mathbf{g}}_n^{i_n}\circ P_{q_n}^n(v_n)-\hat{\mathbf{g}}_1^{i_1}(v_1)\cdots\hat{\mathbf{g}}_n^{i_n}(v_n)\right|\\
    =&M\sum_{i_1,\cdots,i_n}\left|\sum_{k=1}^n\left(\prod_{j=1}^{k-1}\hat{\mathbf{g}}_j^{i_j}\circ P_{q_j}^j(v_j)\cdot\prod_{j=k+1}^n\hat{\mathbf{g}}_j^{i_j}(v_j)\cdot\left(\hat{\mathbf{g}}_k^{i_k}\circ P_{q_k}^k(v_k)-\hat{\mathbf{g}}_k^{i_k}(v_k)\right)\right)\right|\\
    \leq&M\sum_{i_1,\cdots,i_n}\sum_{k=1}^n\left(\prod_{j=1}^{k-1}\hat{\mathbf{g}}_j^{i_j}\circ P_{q_j}^j(v_j)\cdot\prod_{j=k+1}^n\hat{\mathbf{g}}_j^{i_j}(v_j)\cdot\left|\hat{\mathbf{g}}_k^{i_k}\circ P_{q_k}^k(v_k)-\hat{\mathbf{g}}_k^{i_k}(v_k)\right|\right)\\
    =&M\sum_{k=1}^n\sum_{i_1,\cdots,i_n}\left(\prod_{j=1}^{k-1}\hat{\mathbf{g}}_j^{i_j}\circ P_{q_j}^j(v_j)\cdot\prod_{j=k+1}^n\hat{\mathbf{g}}_j^{i_j}(v_j)\cdot\left|\hat{\mathbf{g}}_k^{i_k}\circ P_{q_k}^k(v_k)-\hat{\mathbf{g}}_k^{i_k}(v_k)\right|\right)\\
    \leq&M\sum_{k=1}^n\sum_{i_k}\left|\hat{\mathbf{g}}_k^{i_k}\circ P_{q_k}^k(v_k)-\hat{\mathbf{g}}_k^{i_k}(v_k)\right|\\
    =&M\sum_{k=1}^n\norm{\hat{\mathbf{g}}_k\circ P_{q_k}^k(v_k)-\hat{\mathbf{g}}_k(v_k)}_1. \\
\end{split}
\end{equation*}
Note that $\sum_{i}\hat{\mathbf{g}}_j^i(x)\in [0,1]$ for all $x\in X_j$. Therefore,
\begin{equation*}
\begin{split}
    &\norm{\mathcal{G}(v_1,\cdots,v_n)-\mathbf{u}\langle\hat{\mathbf{g}}_1(P_{q_1}^1(v_1)),\cdots,\hat{\mathbf{g}}_n(P_{q_n}^n(v_n))\rangle} \\
    \leq&\norm{\mathcal{G}(v_1,\cdots,v_n)-\mathbf{u}\langle\hat{\mathbf{g}}_1(v_1),\cdots,\hat{\mathbf{g}}_n(v_n)\rangle}+\\
    &\norm{\mathbf{u}\langle\hat{\mathbf{g}}_1(P_{q_1}^1(v_1)),\cdots,\hat{\mathbf{g}}_n(P_{q_n}^n(v_n))\rangle-\mathbf{u}\langle\hat{\mathbf{g}}_1(v_1),\cdots,\hat{\mathbf{g}}_n(v_n)\rangle}\\
    <&\epsilon+M\sum_{k=1}^n\norm{\hat{\mathbf{g}}_k\circ P_{q_k}^k(v_k)-\hat{\mathbf{g}}_k(v_k)}_1\\
    \leq&\epsilon+M\sum_{k=1}^nL_k^\epsilon(q_k).
\end{split}
\end{equation*}
\qed

\paragraph{Proof of Theorem  \ref{thm:approximation}.}

By Theorem \ref{thm:separate}, for any $\epsilon>0$, there exist positive integers $p_i,q_i$, continuous vector functionals $\hat{\mathbf{g}}_i\in C(X_i,\R^{p_i})$, and $\mathbf{u}\in Y^{p_1\times p_2\times\cdots\times p_n}$, such that
\begin{equation*}
    \sup_{v_i\in K_i}\norm{\mathcal{G}(v_1,\cdots,v_n)-\mathbf{u}\langle\hat{\mathbf{g}}_1(P_{q_1}^1(v_1)),\cdots,\hat{\mathbf{g}}_n(P_{q_n}^n(v_n))\rangle}<\epsilon.
\end{equation*}
Now define
\begin{equation*}
    \mathbf{g}_i=\hat{\mathbf{g}}_i\circ\psi_{q_i}^i,
\end{equation*}
and then we obtain this theorem.
\qed

\paragraph{Proof of Corollary \ref{cor:equiform}.} ~\\
(\ref{eq:approximation})$\Rightarrow$(\ref{eq:elementwise}): Denote $\mathbf{g}_i=(g_j^i)$, and we have
\begin{equation*}
    \mathbf{u}\langle\mathbf{g}_1,\cdots,\mathbf{g}_n\rangle=\sum_{j_1,\cdots ,j_n}g_{j_1}^1\cdots g_{j_n}^nu_{j_1\cdots j_n},
\end{equation*}
which is indeed the form (\ref{eq:elementwise}) by rearrangement and relabeling of the summation.

\noindent(\ref{eq:elementwise})$\Rightarrow$(\ref{eq:symm}): Denote $\mathbf{g}_i=(g_j^i)$, $\mathbf{u}=(u_j)$, and then
\begin{equation*}
\sum_{j=1}^{p}g_j^1\cdots g_j^n u_j=\sum_{j_1,\cdots,j_{n+1}}\delta_{j_1\cdots j_{n+1}}g_{j_{1}}^1\cdots g_{j_n}^nu_{j_{n+1}}=(\delta_{j_1\cdots j_{n+1}})\langle\mathbf{g}_1,\cdots,\mathbf{g}_n,\mathbf{u}\rangle,
\end{equation*}
where $\delta_{j_1\cdots j_{n+1}}$ is equal to 1 if $j_1=\cdots=j_{n+1}$, otherwise it is 0. Moreover, if $u_j$ is approximated by $\tilde{u}_j=\sum_{k=1}^r\alpha_k^je_k$, denote $\tilde{\mathbf{u}}=(\tilde{u}_j)$, $\mathbf{e}=(e_j)$, and then
\begin{equation*}
    (\delta_{j_1\cdots j_{n+1}})\langle\mathbf{g}_1,\cdots,\mathbf{g}_n,\tilde{\mathbf{u}}\rangle=W\langle\mathbf{g}_1,\cdots,\mathbf{g}_n,\mathbf{e}\rangle,
\end{equation*}
where $W=(\sum_{j_{n+1}}\delta_{j_1\cdots j_{n+1}}\alpha_k^{j_{n+1}})$.

\noindent(\ref{eq:symm})$\Rightarrow$(\ref{eq:approximation}): Denote $W=(w_{j_1\cdots j_{n+1}})$, $\mathbf{u}=(u_j)$, so
\begin{equation*}
    W\langle\mathbf{g}_1,\cdots,\mathbf{g}_n,\mathbf{u}\rangle=\tilde{\mathbf{u}}\langle\mathbf{g}_1,\cdots,\mathbf{g}_n\rangle,
\end{equation*}
where $\tilde{\mathbf{u}}=(\sum_{j_{n+1}}w_{j_1\cdots j_{n+1}}u_{j_{n+1}})$.
\qed

\paragraph{Proof of Corollary  \ref{cor:bias}.}

Replace $\mathcal{G}(v_1,\cdots,v_n)$ by $\mathcal{G}(v_1,\cdots,v_n)-b$ in Theorem \ref{thm:separate}, where
\begin{equation*}
    b=\frac{1}{2}\left(\max_{v_i\in K_i,y\in K_0}\mathcal{G}(v_1,\cdots,v_n)(y)+\min_{v_i\in K_i,y\in K_0}\mathcal{G}(v_1,\cdots,v_n)(y)\right).
\end{equation*}
\qed

\paragraph{Proof of Corollary  \ref{cor:multi_output}.}

Replace $Y$ by $Y_1\times\cdots\times Y_m$ in Theorem \ref{thm:approximation}. 
\qed

\paragraph{Proof of Corollary \ref{cor:linear}.}

Without loss of generality, assume that $\mathcal{G}$ is linear with respect to $v_1$, that is, there is a continuous operator defined on $X_1\times K_2\times\cdots\times K_n$ which is linear with respect to $v_1$ and equal to $\mathcal{G}$ limited on $K_1\times\cdots\times K_n$, and for convenience we still denote it as $\mathcal{G}$. Suppose that $\{e_i\},\{e_i^*\}$ are the Schauder basis and coordinate functionals of $X_1$. For $\epsilon>0$, according to the continuity of $\mathcal{G}$ and Property \ref{pro:projection}, there exists a positive integer $q_1$ such that
\begin{equation*}
    \sup_{v_i\in K_i}\norm{\mathcal{G}(v_1,v_2,\cdots,v_n)-\mathcal{G}(P_{q_1}^1(v_1),v_2,\cdots,v_n)}<\frac{\epsilon}{2}.
\end{equation*}
Denote $M=\max_{v_1\in K_1,1\leq j\leq q_1} |e_j^*(v_1)|$. Now define continuous operators $\mathcal{G}_j:K_2\times\cdots\times K_n\to Y$ as
\begin{equation*} \mathcal{G}_j(v_2,\cdots,v_n)=\mathcal{G}(e_j,v_2,\cdots,v_n),\quad 1\leq j\leq q_1.
\end{equation*}
Then by Corollary \ref{cor:multi_output}, there exist positive integers $p_i$, $q_i$, continuous vector functions $\mathbf{g}_i\in C(\R^{q_i},\R^{p_i})$, and $\mathbf{u}_j=(u^j_{k_2\cdots k_n})\in Y^{p_2\times\cdots\times p_n}$, $2\leq i\leq n$, $1\leq j\leq q_1$, such that
\begin{equation*}
    \sup_{v_i\in K_i}\norm{\mathcal{G}_j(v_2,\cdots,v_n)-\mathbf{u}_j\langle\mathbf{g}_2(\varphi_{q_2}^2(v_2)),\cdots,\mathbf{g}_n(\varphi_{q_n}^n(v_n))\rangle}<\frac{\epsilon}{2q_1 M},\quad j=1,\cdots,q_1.
\end{equation*}
Let $p_1=q_1$, $\mathbf{u}=(u_{k_2\cdots k_n}^{k_1})\in Y^{p_1\times\cdots\times p_n}$, $\mathbf{g}_1:\R^{q_1}\to\R^{q_1}$ is the identity map, then
\begin{equation*}
\begin{split}
&\norm{\mathcal{G}(P_{q_1}^1 (v_1),v_2,\cdots,v_n)-\mathbf{u}\langle\mathbf{g}_1(\varphi_{q_1}^1(v_1)),\cdots,\mathbf{g}_n(\varphi_{q_n}^n(v_n))\rangle} \\
=&\norm{\sum_{j=1}^{q_1}e_j^*(v_1)\mathcal{G}(e_j,v_2,\cdots,v_n)-\sum_{j=1}^{q_1}e_j^*(v_1)\mathbf{u}_j\langle\mathbf{g}_2(\varphi_{q_2}^2(v_2)),\cdots,\mathbf{g}_n(\varphi_{q_n}^n(v_n))\rangle} \\
\leq&\sum_{j=1}^{q_1}|e_j^*(v_1)|\cdot\norm{\mathcal{G}(e_j,v_2,\cdots,v_n)-\mathbf{u}_j\langle\mathbf{g}_2(\varphi_{q_2}^2(v_2)),\cdots,\mathbf{g}_n(\varphi_{q_n}^n(v_n))\rangle} \\
<&q_1\cdot M\cdot \frac{\epsilon}{2q_1 M}=\frac{\epsilon}{2}.
\end{split}
\end{equation*}
Therefore,
\begin{equation*}
\sup_{v_i\in K_i}\norm{\mathcal{G}(v_1,\cdots,v_n)-\mathbf{u}\langle\mathbf{g}_1(\varphi_{q_1}^1(v_1)),\cdots,\mathbf{g}_n(\varphi_{q_n}^n(v_n))\rangle}<\frac{\epsilon}{2}+\frac{\epsilon}{2}=\epsilon,
\end{equation*}
where $\mathbf{g}_1$ is linear. The proofs for the other two cases are similar. 
\qed

\paragraph{Proof of Property \ref{pro:rm_lin}.}

Assume that $W_i=(w_j^i)_{p_i}=(w_{jk}^i)_{p_i\times h_i}$, $\mathbf{h}_i=(H_j^i)_{h_i}$, then for $\mathbf{u}=(u_{j_1\cdots j_n})\in Y^{p_1\times p_2\times\cdots\times p_n}$,
\begin{equation*}
\begin{split}
    \mathbf{u}\langle\mathbf{g}_1,\cdots,\mathbf{g}_n\rangle
    =&\sum_{j_1,\cdots,j_n}u_{j_1\cdots j_n}(w_{j_1}^1\mathbf{h}_1)\cdots(w_{j_n}^n\mathbf{h}_n)\\
    =&\sum_{j_1,\cdots,j_n}u_{j_1\cdots j_n}(\sum_{k}w_{j_1k}^1H_k^1)\cdots(\sum_{k}w_{j_nk}^nH_k^n)\\
    =&\sum_{k_1,\cdots,k_n}(\sum_{j_1,\cdots,j_n}u_{j_1\cdots j_n}w_{j_1k_1}^1\cdots w_{j_nk_n}^n)H_{k_1}^1\cdots H_{k_n}^n\\
    =&\tilde{\mathbf{u}}\langle\mathbf{h}_1,\cdots,\mathbf{h}_n\rangle,
\end{split}
\end{equation*}
where $\tilde{\mathbf{u}}=(\sum_{j_1,\cdots,j_n}u_{j_1\cdots j_n}w_{j_1k_1}^1\cdots w_{j_nk_n}^n)_{h_1\times\cdots\times h_n}$. Briefly speaking, the linear output layers of $\mathbf{g}_i$ can be merged into $\mathbf{u}$. The proof for the other case is similar.
\qed

\paragraph{Proof of Corollary \ref{cor:function}.}

We take $X_i=\R^{q_i}$ and $Y=\R$ in Lemma \ref{lem:functional}. Then the corollary can be obtained by the universal approximation theorem for fully-connected neural networks with one hidden layer and Property \ref{pro:rm_lin}.

\section{MIONet for finite-dimensional image space}
\label{app:finite_im}

Corollary \ref{cor:equiform}(i) also gives the approximation theorem for the operators projecting onto both finite-dimensional domain and image space. Given $(v_1,\cdots,v_n,\mathcal{G}(v_1,\cdots,v_n))$, we first transform them to a training set
\begin{equation*}
\{\varphi_{q_1}^1(v_1^k),\cdots,\varphi_{q_n}^n(v_n^k),\varphi_m^{Y}(\mathcal{G}(v_1^k,\cdots,v_n^k))\}_{k=1}^{N},    
\end{equation*}
by determining basis elements $\{e_i\}_{i=1}^{m}$ for $Y$ with $\varphi_m^{Y}(x)=(e_1^*(x),\cdots,e_m^*(x))^T$. Then the loss function can be written as
\begin{equation*}
    \text{MSE} = \frac{1}{mN}\sum_{k=1}^{N}\norm{\varphi_m^{Y}(\mathcal{G}(v_1^k,\cdots,v_n^k))-\left(W_i\langle\tilde{\mathbf{g}}_1(\varphi_{q_1}^1(v^k_1)),\cdots,\tilde{\mathbf{g}}_n(\varphi_{q_n}^n(v^k_n))\rangle\right)_{i=1}^m-b}_2^2,
\end{equation*}
where $\tilde{\mathbf{g}}_i:\R^{q_i}\to\R^{p_i}$ are neural networks to be trained, $W_i\in\R^{p_1\times\cdots\times p_n}$, $b=(b_i)\in\R^m$ are trainable weights and bias respectively. After training, we make prediction by
\begin{equation*}
    \tilde{\mathcal{G}}(v_1,\cdots,v_n) = \sum_{i=1}^m\left(W_i\langle\tilde{\mathbf{g}}_1(\varphi_{q_1}^1(v_1)),\cdots,\tilde{\mathbf{g}}_n(\varphi_{q_n}^n(v_n))\rangle+b_i\right)\cdot e_i.
\end{equation*}

\section{Loss function via numerical integration}\label{app:num_int}
Suppose that $Y=C[0,1]$. For $\mathcal{T}=\{v_1^k,\cdots,v_n^k,\mathcal{G}(v_1^k,\cdots,v_n^k)\}_{k=1}^N$, the general loss function can be computed as
\begin{equation*}
    \mathcal{L}(\mathcal{T})=\frac{1}{N}\sum_{k=1}^N\mathbf{I}(\mathcal{G}(v_1^k,\cdots,v_n^k)-\tilde{\mathcal{G}}(v_1^k,\cdots,v_n^k)),
\end{equation*}
where $\mathbf{I}(\cdot)$ is a numerical integration. For example, for $\mathbf{x}_k$ uniformly sampled on $[0,1]$ ($0=x_0<\cdots<x_m=1$), we have the following choices of $\mathbf{I}(\cdot)$:
\begin{itemize}
    \item rectangle rule: \begin{equation*}
    \mathbf{I}_{rec}(f)=\frac{1}{m}\left(\sum_{k=1}^{m}f\left(\frac{x_{k-1}+x_k}{2}\right)\right),
    \end{equation*}
    \item trapezoidal rule: \begin{equation*}
    \mathbf{I}_{tra}(f)=\frac{1}{m}\left(\sum_{k=1}^{m}\frac{f(x_{k-1})+f(x_k)}{2}\right),
    \end{equation*}
    \item Monte Carlo integration: \begin{equation*}
    \mathbf{I}_{mon}(f)=\frac{1}{m}\left(\sum_{k=1}^{m}f(\mathbf{x}_{k})\right).
    \end{equation*}
\end{itemize}
For a high-dimensional integration, Monte Carlo integration usually performs better.

\bibliographystyle{abbrv}
\bibliography{references} 

\begin{thebibliography}{10}

\bibitem{bhattacharya2020model}
K.~Bhattacharya, B.~Hosseini, N.~B. Kovachki, and A.~M. Stuart.
\newblock Model reduction and neural networks for parametric {PDEs}.
\newblock {\em arXiv preprint arXiv:2005.03180}, 2020.

\bibitem{cai2021deepm}
S.~Cai, Z.~Wang, L.~Lu, T.~A. Zaki, and G.~E. Karniadakis.
\newblock {DeepM\&Mnet}: Inferring the electroconvection multiphysics fields
  based on operator approximation by neural networks.
\newblock {\em Journal of Computational Physics}, 436:110296, 2021.

\bibitem{chen1995universal}
T.~Chen and H.~Chen.
\newblock Universal approximation to nonlinear operators by neural networks
  with arbitrary activation functions and its application to dynamical systems.
\newblock {\em IEEE Transactions on Neural Networks}, 6(4):911--917, 1995.

\bibitem{chen2020physics}
Y.~Chen, L.~Lu, G.~E. Karniadakis, and L.~Dal~Negro.
\newblock Physics-informed neural networks for inverse problems in nano-optics
  and metamaterials.
\newblock {\em Optics Express}, 28(8):11618--11633, 2020.

\bibitem{daneker2022systems}
M.~Daneker, Z.~Zhang, G.~E. Karniadakis, and L.~Lu.
\newblock Systems biology: Identifiability analysis and parameter
  identification via systems-biology informed neural networks.
\newblock {\em arXiv preprint arXiv:2202.01723}, 2022.

\bibitem{deng2021convergence}
B.~Deng, Y.~Shin, L.~Lu, Z.~Zhang, and G.~E. Karniadakis.
\newblock Convergence rate of {DeepONets} for learning operators arising from
  advection-diffusion equations.
\newblock {\em arXiv preprint arXiv:2102.10621}, 2021.

\bibitem{di2021deeponet}
P.~C. Di~Leoni, L.~Lu, C.~Meneveau, G.~Karniadakis, and T.~A. Zaki.
\newblock {DeepONet} prediction of linear instability waves in high-speed
  boundary layers.
\newblock {\em arXiv preprint arXiv:2105.08697}, 2021.

\bibitem{fabian2011banach}
M.~Fabian, P.~Habala, P.~H{\'a}jek, V.~Montesinos, and V.~Zizler.
\newblock {\em Banach space theory: the basis for linear and nonlinear
  analysis}.
\newblock Springer Science \& Business Media, 2011.

\bibitem{goswami2022physics}
S.~Goswami, M.~Yin, Y.~Yu, and G.~E. Karniadakis.
\newblock A physics-informed variational {DeepONet} for predicting crack path
  in quasi-brittle materials.
\newblock {\em Computer Methods in Applied Mechanics and Engineering},
  391:114587, 2022.

\bibitem{johan1990tensor}
H.~Johan.
\newblock Tensor rank is {NP}-complete.
\newblock {\em Journal of Algorithms}, 4(11):644--654, 1990.

\bibitem{karniadakis2021physics}
G.~E. Karniadakis, I.~G. Kevrekidis, L.~Lu, P.~Perdikaris, S.~Wang, and
  L.~Yang.
\newblock Physics-informed machine learning.
\newblock {\em Nature Reviews Physics}, 3(6):422--440, 2021.

\bibitem{adam2015}
D.~P. Kingma and J.~Ba.
\newblock Adam: {A} method for stochastic optimization.
\newblock In {\em 3rd International Conference on Learning Representations,
  {ICLR} 2015, San Diego, CA, USA, May 7-9, 2015, Conference Track
  Proceedings}, 2015.

\bibitem{kissas2022learning}
G.~Kissas, J.~Seidman, L.~F. Guilhoto, V.~M. Preciado, G.~J. Pappas, and
  P.~Perdikaris.
\newblock Learning operators with coupled attention.
\newblock {\em arXiv preprint arXiv:2201.01032}, 2022.

\bibitem{kissas2020machine}
G.~Kissas, Y.~Yang, E.~Hwuang, W.~R. Witschey, J.~A. Detre, and P.~Perdikaris.
\newblock Machine learning in cardiovascular flows modeling: Predicting
  arterial blood pressure from non-invasive {4D} flow {MRI} data using
  physics-informed neural networks.
\newblock {\em Computer Methods in Applied Mechanics and Engineering},
  358:112623, 2020.

\bibitem{kolda2009tensor}
T.~G. Kolda and B.~W. Bader.
\newblock Tensor decompositions and applications.
\newblock {\em SIAM Review}, 51(3):455--500, 2009.

\bibitem{kovachki2021universal}
N.~Kovachki, S.~Lanthaler, and S.~Mishra.
\newblock On universal approximation and error bounds for {Fourier} neural
  operators.
\newblock {\em Journal of Machine Learning Research}, 22:Art--No, 2021.

\bibitem{kovachki2021neural}
N.~Kovachki, Z.~Li, B.~Liu, K.~Azizzadenesheli, K.~Bhattacharya, A.~Stuart, and
  A.~Anandkumar.
\newblock Neural operator: Learning maps between function spaces.
\newblock {\em arXiv preprint arXiv:2108.08481}, 2021.

\bibitem{kruskal1989rank}
J.~B. Kruskal.
\newblock Rank, decomposition, and uniqueness for 3-way and n-way arrays.
\newblock {\em Multiway data analysis}, pages 7--18, 1989.

\bibitem{lanthaler2021error}
S.~Lanthaler, S.~Mishra, and G.~E. Karniadakis.
\newblock Error estimates for {DeepONets}: A deep learning framework in
  infinite dimensions.
\newblock {\em arXiv preprint arXiv:2102.09618}, 2021.

\bibitem{li2020fourier}
Z.~Li, N.~Kovachki, K.~Azizzadenesheli, B.~Liu, K.~Bhattacharya, A.~Stuart, and
  A.~Anandkumar.
\newblock Fourier neural operator for parametric partial differential
  equations.
\newblock {\em arXiv preprint arXiv:2010.08895}, 2020.

\bibitem{li2020neural}
Z.~Li, N.~Kovachki, K.~Azizzadenesheli, B.~Liu, K.~Bhattacharya, A.~Stuart, and
  A.~Anandkumar.
\newblock Neural operator: Graph kernel network for partial differential
  equations.
\newblock {\em arXiv preprint arXiv:2003.03485}, 2020.

\bibitem{lin2021operator}
C.~Lin, Z.~Li, L.~Lu, S.~Cai, M.~Maxey, and G.~E. Karniadakis.
\newblock Operator learning for predicting multiscale bubble growth dynamics.
\newblock {\em The Journal of Chemical Physics}, 154(10):104118, 2021.

\bibitem{lin2021seamless}
C.~Lin, M.~Maxey, Z.~Li, and G.~E. Karniadakis.
\newblock A seamless multiscale operator neural network for inferring bubble
  dynamics.
\newblock {\em Journal of Fluid Mechanics}, 929, 2021.

\bibitem{lin2021accelerated}
G.~Lin, C.~Moya, and Z.~Zhang.
\newblock Accelerated replica exchange stochastic gradient {Langevin} diffusion
  enhanced {Bayesian DeepONet} for solving noisy parametric {PDEs}.
\newblock {\em arXiv preprint arXiv:2111.02484}, 2021.

\bibitem{liu2021multiscale}
L.~Liu and W.~Cai.
\newblock Multiscale {DeepONet} for nonlinear operators in oscillatory function
  spaces for building seismic wave responses.
\newblock {\em arXiv preprint arXiv:2111.04860}, 2021.

\bibitem{lu2019deeponet}
L.~Lu, P.~Jin, and G.~E. Karniadakis.
\newblock {DeepONet}: Learning nonlinear operators for identifying differential
  equations based on the universal approximation theorem of operators.
\newblock {\em arXiv preprint arXiv:1910.03193}, 2019.

\bibitem{lu2021learning}
L.~Lu, P.~Jin, G.~Pang, Z.~Zhang, and G.~E. Karniadakis.
\newblock Learning nonlinear operators via {DeepONet} based on the universal
  approximation theorem of operators.
\newblock {\em Nature Machine Intelligence}, 3(3):218--229, 2021.

\bibitem{lu2021comprehensive}
L.~Lu, X.~Meng, S.~Cai, Z.~Mao, S.~Goswami, Z.~Zhang, and G.~E. Karniadakis.
\newblock A comprehensive and fair comparison of two neural operators (with
  practical extensions) based on fair data.
\newblock {\em arXiv preprint arXiv:2111.05512}, 2021.

\bibitem{lu2021deepxde}
L.~Lu, X.~Meng, Z.~Mao, and G.~E. Karniadakis.
\newblock {DeepXDE}: A deep learning library for solving differential
  equations.
\newblock {\em SIAM Review}, 63(1):208--228, 2021.

\bibitem{lu2021physics}
L.~Lu, R.~Pestourie, W.~Yao, Z.~Wang, F.~Verdugo, and S.~G. Johnson.
\newblock Physics-informed neural networks with hard constraints for inverse
  design.
\newblock {\em SIAM Journal on Scientific Computing}, 43(6):B1105--B1132, 2021.

\bibitem{mao2021deepm}
Z.~Mao, L.~Lu, O.~Marxen, T.~A. Zaki, and G.~E. Karniadakis.
\newblock {DeepM\&Mnet} for hypersonics: Predicting the coupled flow and
  finite-rate chemistry behind a normal shock using neural-network
  approximation of operators.
\newblock {\em Journal of Computational Physics}, 447:110698, 2021.

\bibitem{marcati2021exponential}
C.~Marcati and C.~Schwab.
\newblock Exponential convergence of deep operator networks for elliptic
  partial differential equations.
\newblock {\em arXiv preprint arXiv:2112.08125}, 2021.

\bibitem{nelsen2021random}
N.~H. Nelsen and A.~M. Stuart.
\newblock The random feature model for input-output maps between {Banach}
  spaces.
\newblock {\em SIAM Journal on Scientific Computing}, 43(5):A3212--A3243, 2021.

\bibitem{osorio2022forecasting}
J.~D. Osorio, Z.~Wang, G.~Karniadakis, S.~Cai, C.~Chryssostomidis, M.~Panwar,
  and R.~Hovsapian.
\newblock Forecasting solar-thermal systems performance under transient
  operation using a data-driven machine learning approach based on the deep
  operator network architecture.
\newblock {\em Energy Conversion and Management}, 252:115063, 2022.

\bibitem{pang2019fpinns}
G.~Pang, L.~Lu, and G.~E. Karniadakis.
\newblock {fPINNs}: Fractional physics-informed neural networks.
\newblock {\em SIAM Journal on Scientific Computing}, 41(4):A2603--A2626, 2019.

\bibitem{patel2021physics}
R.~G. Patel, N.~A. Trask, M.~A. Wood, and E.~C. Cyr.
\newblock A physics-informed operator regression framework for extracting
  data-driven continuum models.
\newblock {\em Computer Methods in Applied Mechanics and Engineering},
  373:113500, 2021.

\bibitem{raissi2019physics}
M.~Raissi, P.~Perdikaris, and G.~E. Karniadakis.
\newblock Physics-informed neural networks: A deep learning framework for
  solving forward and inverse problems involving nonlinear partial differential
  equations.
\newblock {\em Journal of Computational Physics}, 378:686--707, 2019.

\bibitem{raissi2020hidden}
M.~Raissi, A.~Yazdani, and G.~E. Karniadakis.
\newblock Hidden fluid mechanics: Learning velocity and pressure fields from
  flow visualizations.
\newblock {\em Science}, 367(6481):1026--1030, 2020.

\bibitem{ryan2002introduction}
R.~A. Ryan.
\newblock {\em Introduction to tensor products of {Banach} spaces}, volume~73.
\newblock Springer, 2002.

\bibitem{sirignano2018dgm}
J.~Sirignano and K.~Spiliopoulos.
\newblock {DGM}: A deep learning algorithm for solving partial differential
  equations.
\newblock {\em Journal of computational physics}, 375:1339--1364, 2018.

\bibitem{trask2019gmls}
N.~Trask, R.~G. Patel, B.~J. Gross, and P.~J. Atzberger.
\newblock {GMLS-Nets}: A framework for learning from unstructured data.
\newblock {\em arXiv preprint arXiv:1909.05371}, 2019.

\bibitem{wang2021learning}
S.~Wang, H.~Wang, and P.~Perdikaris.
\newblock Learning the solution operator of parametric partial differential
  equations with physics-informed {DeepONets}.
\newblock {\em Science Advances}, 7(40):eabi8605, 2021.

\bibitem{weinan2018deep}
E.~Weinan and B.~Yu.
\newblock The deep {Ritz} method: A deep learning-based numerical algorithm for
  solving variational problems.
\newblock {\em Communications in Mathematics and Statistics}, 6(1), 2018.

\bibitem{yazdani2020systems}
A.~Yazdani, L.~Lu, M.~Raissi, and G.~E. Karniadakis.
\newblock Systems biology informed deep learning for inferring parameters and
  hidden dynamics.
\newblock {\em PLoS Computational Biology}, 16(11):e1007575, 2020.

\bibitem{yin2022simulating}
M.~Yin, E.~Ban, B.~V. Rego, E.~Zhang, C.~Cavinato, J.~D. Humphrey, and
  G.~Em~Karniadakis.
\newblock Simulating progressive intramural damage leading to aortic dissection
  using {DeepONet}: an operator--regression neural network.
\newblock {\em Journal of the Royal Society Interface}, 19(187):20210670, 2022.

\bibitem{you2022nonlocal}
H.~You, Y.~Yu, M.~D'Elia, T.~Gao, and S.~Silling.
\newblock Nonlocal kernel network ({NKN}): a stable and resolution-independent
  deep neural network.
\newblock {\em arXiv preprint arXiv:2201.02217}, 2022.

\bibitem{yu2021gradient}
J.~Yu, L.~Lu, X.~Meng, and G.~E. Karniadakis.
\newblock Gradient-enhanced physics-informed neural networks for forward and
  inverse {PDE} problems.
\newblock {\em arXiv preprint arXiv:2111.02801}, 2021.

\bibitem{zhang2019quantifying}
D.~Zhang, L.~Lu, L.~Guo, and G.~E. Karniadakis.
\newblock Quantifying total uncertainty in physics-informed neural networks for
  solving forward and inverse stochastic problems.
\newblock {\em Journal of Computational Physics}, 397:108850, 2019.

\end{thebibliography}

\end{document}